\documentclass[lettersize,journal]{IEEEtran}
\usepackage{amsmath,amsfonts}
\usepackage{algorithmic}
\usepackage{algorithm}
\usepackage{array}
\usepackage[caption=false,font=normalsize,labelfont=sf,textfont=sf]{subfig}
\usepackage{textcomp}
\usepackage{stfloats}
\usepackage{bibentry}
\usepackage{url}
\usepackage{verbatim}
\usepackage{graphicx}
\usepackage{cite}
\usepackage{multirow}
\usepackage{balance}
\usepackage{subfloat}
\usepackage{threeparttable}
\usepackage{makecell}
\usepackage{stfloats}

\def\BibTeX{{\rm B\kern-.05em{\sc i\kern-.025em b}\kern-.08em
		T\kern-.1667em\lower.7ex\hbox{E}\kern-.125emX}}
\usepackage{balance}
\begin{document}
	\title{Vision Aided Channel Prediction for Vehicular Communications: A Case Study  of Received Power Prediction Using RGB Images}
	\author{Xuejian Zhang, ~\IEEEmembership{Student Member,~IEEE,} Ruisi He,~\IEEEmembership{Senior Member,~IEEE,} Mi Yang,~\IEEEmembership{Member,~IEEE,} \\ 
		Zhengyu Zhang,~\IEEEmembership{Student Member,~IEEE,} Ziyi Qi, Bo Ai,~\IEEEmembership{Fellow,~IEEE} 
	\thanks{
Part of this paper has been presented in the 
IEEE International Symposium on Antennas and Propagation and ITNC-USNC-URSI Radio Science Meeting (IEEE AP-S-2024) \cite{APS2024}.

X. Zhang, R. He,   Z. Zhang, Z. Qi and B. Ai are with the School of Electronics and Information Engineering and the Frontiers Science Center for Smart High-speed Railway System, Beijing Jiaotong University, Beijing 100044, China
(email: 23115029@bjtu.edu.cn; ruisi.he@bjtu.edu.cn; 21111040@bjtu.edu.cn;  22115006@bjtu.edu.cn;  boai@bjtu.edu.cn).

M. Yang is with the School of Electronics and Information Engineering and the Frontiers Science Center for Smart High-speed Railway System, Beijing Jiaotong University, Beijing 100044, China, and also with the Henan High-Speed Railway Operation and Maintenance Engineering Research Center, Zhengzhou 451460, China (e-mail: myang@bjtu.edu.cn).

}
}
	\maketitle

\begin{abstract}
The communication scenarios and channel characteristics of 6G will be more complex and difficult to characterize.
Conventional methods for channel prediction face challenges in achieving an optimal balance between accuracy, practicality, and generalizability. Additionally, they often fail to effectively leverage environmental features.
Within the framework of integration communication and artificial intelligence as a pivotal development vision for 6G, it is imperative to achieve intelligent prediction of channel characteristics.
Vision-aided methods have been employed in various wireless communication tasks, excluding channel prediction, and have demonstrated enhanced efficiency and performance.
In this paper,
we   propose a  vision-aided two-stage model  for  channel prediction in millimeter wave vehicular communication scenarios, realizing accurate received power prediction   utilizing solely RGB images.
Firstly, 
we obtain  original images of  propagation environment through an RGB camera.
Secondly, 
three typical computer vision methods including object detection, instance segmentation and binary mask are employed for environmental  information extraction from original images in stage 1, and prediction of received power based on processed images is implemented in stage 2.
Pre-trained YOLOv8 and ResNets  are used in stages 1 and 2, respectively,  and  fine-tuned on  datasets.
Finally, we conduct five experiments to evaluate the performance of   proposed model, demonstrating its feasibility, accuracy  and generalization capabilities. 
The model proposed in this paper offers novel solutions for achieving intelligent channel prediction in vehicular communications.

\end{abstract}


\section{Introduction}

  \IEEEPARstart{T}{he} 
future 6G vision articulated by ITU-R highlights the integration of communication and artificial intelligence (AI) as a critical emerging technology trend \cite{itu1,itu2}. 
 Consequently, the development of more efficient and reliable wireless communication systems leveraging AI technology has garnered significant attention in the research community  \cite{Huang2022}.
At the same time,  communication scenarios of 6G are increasingly complex, featuring three-dimensional communication across air-space-ground, higher frequency bands such as millimeter waves and terahertz, wider bandwidths, and massive multiple-input multiple-output (MIMO) systems \cite{hrs2025}. 
These aspects represent both new development trends and  challenges \cite{Graff2023}. 
They heavily utilize valuable spatio-temporal-frequency resources, imposing additional communication costs and greater computational resource consumption on wireless communication systems \cite{Yuxin2023}, and are not conducive to   establishment of wireless communication systems with efficient resource utilization for 6G.

Recently, with explosive advances in computer vision (CV) technology and growing interest in using AI to solve key wireless communication challenges \cite{huangchen,Qiu2024,yangmi}, large amounts of out-of-band information and data, such as visual sensing data, have been introduced into wireless communication systems, since their acquisition does not occupy other frequency resources, it is  beneficial to reduce communication overhead and improve spectrum efficiency. 
What's more, with  gradual development of intelligent services such as intelligent transportation and autonomous driving, vehicle and infrastructure will be installed with  rich variety of sensors such as radar, LiDAR, depth/RGB cameras, which will make  acquisition of visual data  easier \cite{QI2023241}. 
Visual data obtained can be used to assist various vehicular communications, such as vehicle-to-vehicle (V2V) and vehicle-to-infrastructure (V2I) communications, i.e., enabling vision-aided wireless communications \cite{Yang2021}.
It has been shown to significantly enhance wireless communication reliability without sacrificing spectral efficiency \cite{tian,Nishio2021}.

Channel prediction is one of potential typical applications that can be aided by  visual information \cite{he2019applications}. 
Accurate channel prediction is the foundation of intelligent vehicular communications \cite{MiYang2023, hrs2020}.
Moreover, there is actually a natural match between  channel and  visual data because
wireless channel is determined by propagation environment which can be recorded in the form of visual data. 
Based on vision information, sizes, shapes and positions of scatterers  can be reflected, effectively indicating wireless propagation features, and  typical channel characteristics, including received power, path loss, etc.,  can be inferred.
Further, this method  can be used to  provide valuable guidance for  design and deployment of intelligent vehicular communication systems. 
For example, it can assist in determining the optimal number and spacing of base stations, as well as dynamically adjusting base station power in real time to ensure reliable V2V and V2I communications.

To achieve  the above goals, environment information must be  acquired firstly, i.e. environment  sensing. 
RGB images taken by  cameras are   widely used forms of  visual sensing data that contain information about propagation environment due to convenience and low cost \cite{cje}.
Secondly, features of key scatterers are extracted from original visual   data  based on CV.
Finally,  AI-based methods can be used to achieve channel prediction (such as received power prediction) through processed image.
  

\subsection{Related Work}
Classical methods have been widely used in channel prediction.
Narrowband and wideband channel measurements are carried out in  \cite{Zhang2024,Zhang2023,He2013} and   \cite{Zhengyu2023,YiZeng2020,MiYang2020}, respectively. 
Channel characteristics such as path loss (PL),  delay spread, angle spread etc.   are extracted, modeled and analyzed based on   measured data. 
Refs. \cite{Wang2018,Longhe2019}  use  ray tracing to virtually reconstruct  real propagation scenario, and precisely simulates  propagation law of each multipath component to achieve accurate characterization of channel. 
Furthermore, a pervasive channel modeling theory is pioneered in \cite{Wang2022} and a 6G pervasive geometry-based stochastic model (GBSM) is then proposed, which for the first time can model channel characteristics of all frequency bands and all scenarios including mmWave and V2V channels. The statistical properties of   proposed 6G pervasive GBSM are fully investigated in \cite{Wang2023}, providing interesting relationships of model parameters, statistic properties, frequency bands, and scenarios.
The above classical methods usually rely on complex measurement data or high computational costs, and they fail to consider accuracy, complexity and utilization of environment information of at the same time.

The development of CV technology provides a powerful means for processing visual data, enabling vision-aided wireless communications to be implemented in a variety of application scenarios. 
Refs. \cite{alrabeiah2020viwi,Alrabeiah2020,Charan2021} obtain   simulated time series RGB images from   base station and use  recurrent neural networks (RNN) and convolutional neural networks (CNN) respectively to implement beam prediction, blocking prediction and proactive handoff in mmWave MIMO networks.
Refs. \cite{Nishio2019,Koda2020}  have proposed  novel mechanisms for predicting time series of received power and blockage prediction by using convolutional long short-term memory  with depth images. 
Refs. \cite{Xu2023_2,Xu2023,Feng2024} respectively use vision-aided methods with RGB images to achieve beam alignment in mmWave vehicle communications, multi-user matching and resource allocation, as well as ultra-reliable and low-latency communications in smart factory scenarios.
Refs. \cite{env,feifei2023, feifeigao2023} put forward  different concepts of environment semantic to reduce redundancy of images and efficiently achieve  beam and blockage prediction better  in mmWave MIMO networks
It can be seen that these studies all focus on solving problems such as beam prediction and selection, blockage prediction, etc.. 
However, they are not applied to channel prediction to assist  vehicular communications.

To sum up, this work is rarely mentioned in existing research, which is to   realize   accurate and efficient channel prediction based on environment information extraction using vision-aided methods  in  vehicular communications. 
It will be  helpful to promote   further deep integration of AI  and   communication and also provide a set of solutions for     realization of intelligent deployment and application of   vehicular communication system.
  
\subsection{Contributions}
To address  the issues discussed above, we presented some preliminary experimental results in our previous work of \cite{APS2024}.
In this paper, a two-stage deep learning model for channel prediction is proposed for mmWave vehicular communications.
Three mature CV methods and two widely used deep neural networks, i.e. YOLOv8 and ResNets,  are combined.
We believe that this is an interesting application that we can accurately predict channel characteristics, such as received power, using only environment RGB images.
The main contributions of this paper are summarized as follows.

\begin{itemize}
	\item[$\bullet$]  A two-stage channel prediction model for mmWave V2I scenarios based on deep learning is proposed, and detailed steps for training the model using RGB images and wireless channel data are described.
	\item[$\bullet$]  To accurately predict  received power using only RGB images, three typical CV methods are  presented to extract environment  information, which are adding bounding box, binary mask and instance segmentation to  scatterers of interest in original images respectively.
	\item[$\bullet$]  Extensive experiments are carried out based on open source measurement datasets, 
	comparison of  predictive performance of models with different processed images  and with or without interference elimination are used as input,
	generalization of trained model on datasets in different scenarios and conditions, 
	impact of different  neural network scales on model prediction accuracy all have been discussed in detail.
	Results  has validated the excellent performance of   proposed model in practicability, reliability, and generalization.

\end{itemize}

The remainder of this paper is organized as follows. 
Section II describes the framework of two-stage model based on DL proposed in this paper. 
Section III   introduces   experiment  setup and  acquisition, selection and utilizability of datasets. 
Then in Section IV, a series of experiments about performance validation and analysis for   proposed model is presented. 
Finally, Section V draws the conclusions.

 \begin{figure*}[!t]
	\centering
	\includegraphics[width=.95\textwidth]{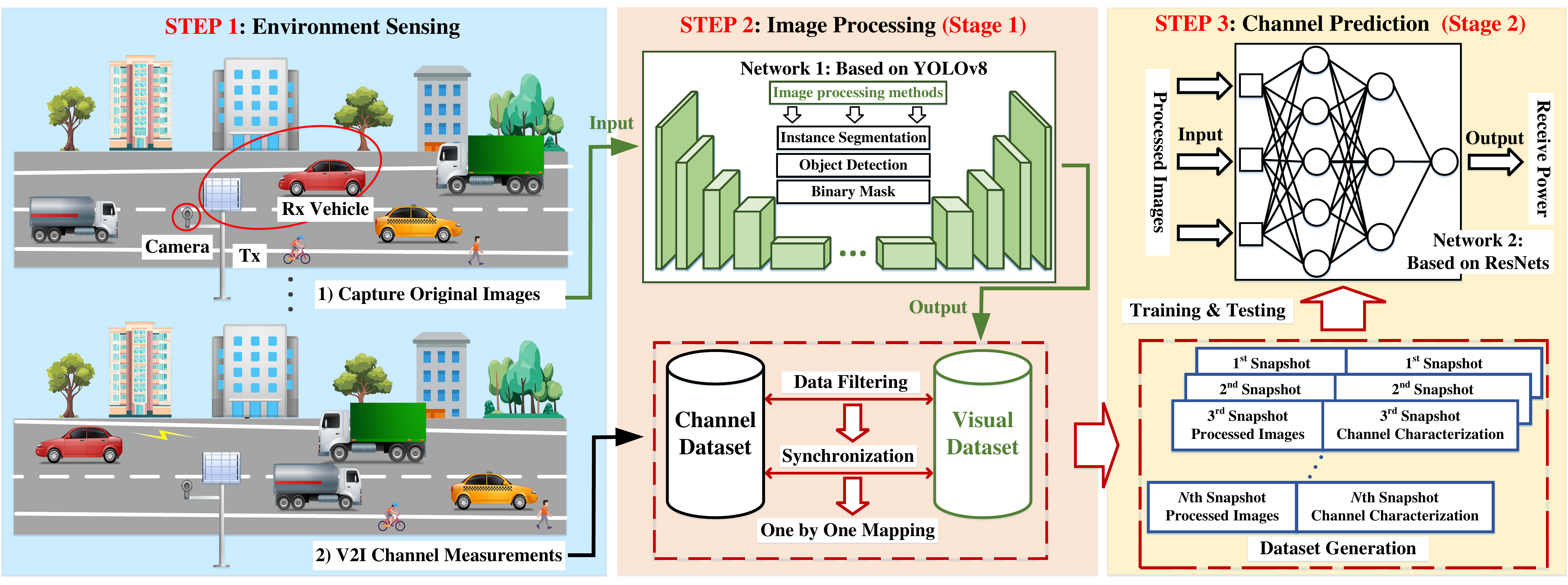}%
	\caption{Framework of proposed two-stage channel prediction model.}                    
	\label{model}
\end{figure*}

\section{Vision-Aid Channel Prediction Model}  
\subsection{System Model}
In this study, we consider a typical  V2I  communication scenario, where a base station (BS) is located at an intersection along the roadside, as shown in Fig. \ref{model}. 
Furthermore, BS is equipped with a transmitting antenna and a standard-resolution RGB wide-angle camera, which together enable communication and environmental monitoring. 
The mobile user (i.e., target vehicle) is equipped with a receiving antenna capable of receiving  vector channel sounding signals transmitted by BS. 
Without considering  MIMO,  channel impulse response $h(t,\tau )$ can be expressed as \cite{myy}
\begin{equation}
	\label{math1}
	h(t,\tau ) = \sum\limits_{l = 1}^L {{\alpha _l}{e^{ - j\phi {}_l}}\delta (t - {\tau _l})}, 
\end{equation}
where $t$ is  index of  time snapshot, $ L$ are the number  of   rays, and $\tau _l $ is the delay of the $ L$-th path. 
What's more, $ \delta ( \cdot )$ is the Dirac delta function, and $ \phi {}_l $ is the phase of paths that is assumed to be described by statistically independent random variables uniformly distributed over $[0,2\pi )$.
Performing Fourier transform on $h(t,\tau )$ can obtain the channel transfer function $H(f)$.
Thus we can obtain  channel characteristics, typical examples of which include PL,  delay spread ${\tau _{rms}}$, received power ${P}$, etc.

\subsection{Problem Formulation}
The task of channel prediction is to capture the evolving trends of channel  in order to inform key aspects of communication systems, such as network planning, base station placement, power control, etc. 
Traditional approaches typically rely on channel models, statistical methods based on channel measurements, or deterministic methods such as ray tracing and parabolic equations, to predict channel propagation characteristics. 
However,  channel models established based on traditional methods often struggle to balance accuracy with computational complexity, lack generalization and adaptability across diverse scenarios, 
and fail to  leverage the characteristics of  propagation environment.

A promising alternative is the application of AI, particularly CV, to perceive  propagation environment and thereby assisting in  channel prediction. 
Visual data, which can be easily acquired and does not consume valuable spectrum resources, represents a valuable out-of-band resource. 
In recent years, research on exploring the use of visual data to assist wireless communication has gained traction, particularly in areas such as blockage prediction, beamforming prediction, and  proactive handoff in mmWave networks.
In this paper,  we propose the use of additional visual data—specifically, RGB images captured by camera installed on BS conjunction with CV and deep learning to predict channel.
Notely, instead of using   raw images, we follow the suggestion of \cite{env} and use RGB images processed by different CV methods.

Formally, we define $ {\rm X}[t] \in \mathbb{R} {^{W \times H \times C}} $ as the corresponding RGB image  at time $t$, where $W$, $H$, $C$ are the width, height, and the number of color channels of the image.
Let, $ {\rm Z}[t]$ represent processed RGB images. 
The objective of channel  prediction task in this paper is to find a mapping function ${f_\Theta }$ that utilizes $ {\rm Z}[t]$ to predict channel characteristics $\widehat c[t] \in \Omega $,  $\Omega  = [PL,{\tau _{rms}},P...]$. 
The mapping function can be formally expressed as
\begin{equation}
	\label{math6}
	{f_\Theta }:{\rm Z}[t] \to \widehat c[t].
\end{equation}

In this paper, we develop a deep learning model to learn this prediction function ${h_\Theta }$.
Let $D = \left\{ {\left( {{Z_m},c_m^ \star } \right)} \right\}_m^M$ represent available dataset consisting of ``processed RGB image-channel characteristic'' pairs is collected from wireless propagation environment as shown in Fig. \ref{model}, where $M$ is the total number of snapshots in the dataset.
In addition,  loss function can be expressed as
\begin{equation}
	\label{math7}
	L = \frac{1}{M}\sum\limits_{m = 1}^M {\zeta (c_m^ \star ,\widehat {{c_m}})} ,
\end{equation}
where $\zeta ( \cdot )$ is the loss of a single snapshot, which measures the difference between   predicted value and  actual value.
Then, the goal is   to minimize the loss over all snapshots in the dataset $D$, which can be formally written as
\begin{equation}
	\label{math8}
	f_{{\Theta ^ \star }}^ \star  = \mathop {\arg \min }\limits_{{f_\Theta }} L(c_m^ \star |{{\rm Z}_m},\widehat {{c_m}}).
\end{equation}
The prediction function is parameterized by a set of model parameters $\Theta$ and is learned from the labeled data  in the dataset $D$. 
The objective is to find the best parameters $\Theta ^ \star$ that minimize the loss function. 
Next, we present the proposed vision based solution for channel   prediction.



 \begin{figure*}[!t]
	\centering
	\includegraphics[width=.90\textwidth]{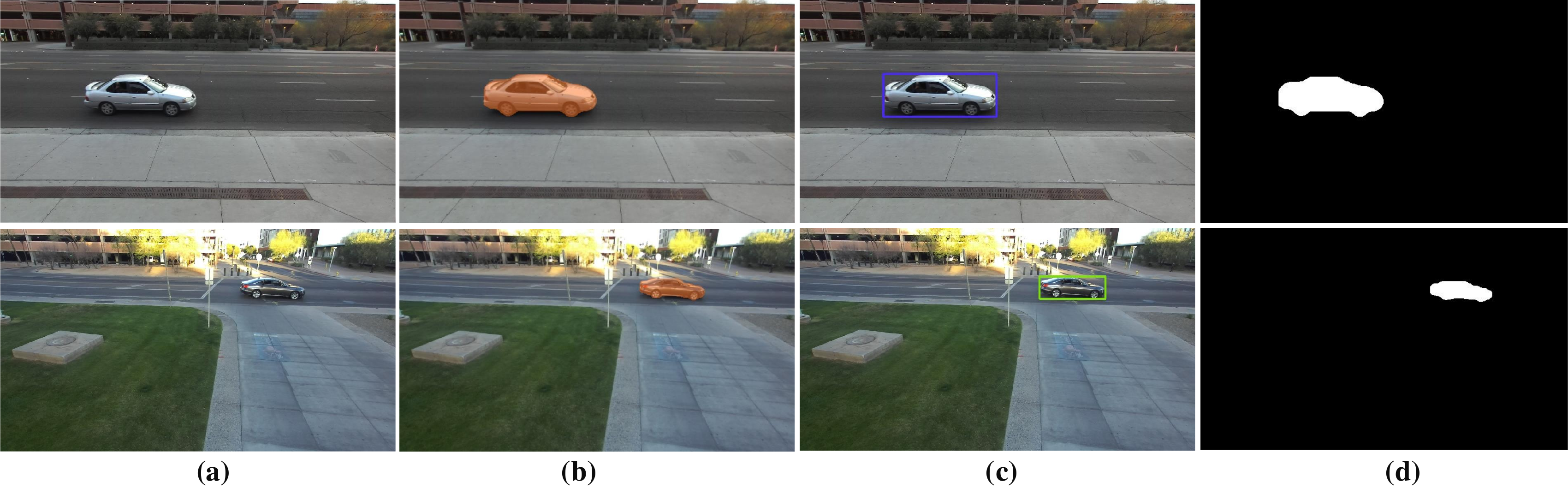}%
	\caption{Examples of three kind of images based on different CV methods.
		(a) Original images;
		(b) Images with instance segmentation;
		(c) Images with object detection;
		(d) Images with binary mask.}                    
	\label{exam}
\end{figure*}

\subsection{Framework}

In this paper, we present an vision-aided two-stage deep learning model to  predict  channel characteristics, especially using received power as a study case. 
Framework of the model  is shown in Fig. \ref{model}, which can be divided into three steps: environment sensing, image processing and channel prediction. 
The complete implementation is described in detail below.

\textbf{Step 1: Environment Sensing.}
In this step, two independent datasets need to be collected. 
First, the camera mounted on BS (or roadside infrastructure) continuously captures RGB images of  propagation environment, forming a original image dataset. 
Second, V2I channel measurements are conducted in the same environment to obtain vector channel data, which constitutes a channel dataset.

\textbf{Step 2: Image processing.}
Image processing involves explicitly labeling target vehicle in the original images using various CV techniques as shown in  Fig. \ref{exam}, which includes object recognition with rectangular bounding boxes, instance segmentation with color masks, or segmentation using black-and-white masks. 
This  facilitates the prediction network in Step 3 by enabling it to more effectively learn image features, thereby improving the accuracy of received power predictions. 
In Fig. \ref{exam}(b),
Outline of target  vehicle is covered by pixels of the same color, and  other scatterers  remain unchanged. 
In Fig. \ref{exam}(c), we use object detection to frame  target vehicle  with a rectangular frame, and other scatterers  also remain unchanged.
Entire image is covered by a binary mask in Fig. \ref{exam}(d), in which  target vehicle is filled with white pixels, while   others are completely filled with black pixels. 
The impact of  three types of images on prediction accuracy will be discussed in detail in Section IV.

In Step 2, original image dataset is firstly fed into Network 1, and processed images are generated to form visual dataset.
The next critical step is to filter and synchronize  channel data with  processed visual data to establish a complete one-to-one mapping.
This is essential because not all   raw data obtained can be directly utilized for training and testing. 
Redundant data, such as images where  target vehicle is absent, or visual and channel data that cannot be synchronized by timestamps due to inconsistent acquisition rates, offer limited value for training prediction network in Step 3.

It should be mentioned that Network 1 in Step 2 we have used  is YOLOv8. 
YOLOv8 \cite{Jocher_Ultralytics_YOLO_2023} is used to implement  extraction of environment  information in this paper.
YOLOv8 is released by  Ultralytics company in 2023, which has shown significant improvements in accuracy and speed compared with previous versions \cite{talaat2023improved}.
More importantly, YOLOv8 has the functions of object detection, instance segmentation and binary mask addition. 
And  pre-trained models provided by Ultralytics company after training on  COCO dataset \cite{COCO}  are also available.
It provides great convenience for extracting  environmental information based on transfer learning.

\textbf{Step 3: Channel Prediction.}
Processed images  are fed into  Network 2 to   output predicted values of channel  characteristics in Step 3, and we take  received power as an example. 
Residual neural network (ResNet) has been widely used in CV tasks  and has  excellent ability to deal with both classification and regression problems.
Therefore,  ResNets proposed in  \cite{he2015deep} with various network scales are adopted and customized to fit received power prediction problem, including ResNet-18, ResNet-34, ResNet-50, ResNet-101 and ResNet-152. 
ResNets we use have been pre-trained on    ImageNet2012 dataset \cite{russakovsky2015imagenet},
and network structure of them are the same as in \cite{he2015deep},  
except for   their final fully connected layers, which are removed and replaced with another fully connected layers of one output. 
This is because   received power prediction is a regression task and  final output is unique power value in dBm.
ResNets are then fine-tuned, in a supervised fashion,  using images from   environment that are labeled with   corresponding received power.
The performance in different scenarios and  impact of different network scales on  prediction accuracy will be discussed in detail in Section IV.


 \begin{figure*}[!t]
	\centering
	\includegraphics[width=.99\textwidth]{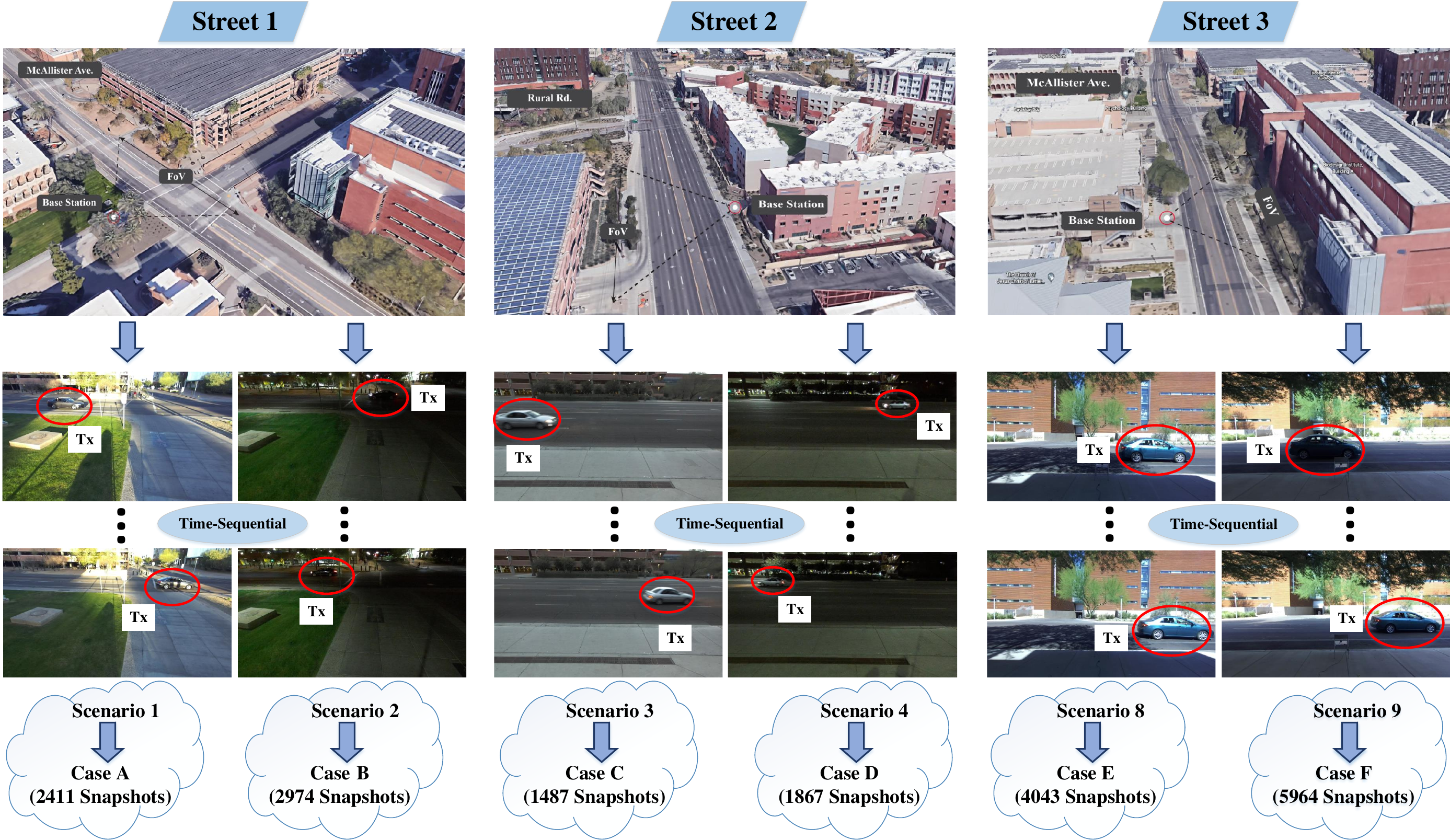}%
	\caption{Measurement scenarios for different streets and different situations (day or night).}                    
	\label{Scenarios}
\end{figure*}

\section{Experimental Setup}
\subsection{Datasets acquisition}
The data used in this paper are from well-known   open source dataset \emph{DeepSense 6G} \cite{deepsense,Charan_DeepSense_2022a,AhmedAlkhateeb2022}.
\emph{DeepSense 6G} is a real-world multi-modal dataset designed to facilitate the development of sensing-aided
wireless communication applications \cite{Charan2024}.
In this paper, only datasets of RGB images and received power have been used and they are derived from  Scenarios 1-4 and 8-9 in \emph{DeepSense 6G}, which are obtained from three different streets (named street 1, street 2 and street 3).
It should be noted that Scenarios 1, 2 and 3, 4 represent both day and night conditions on streets 1 and 2, respectively, and Scenarios 8 and 9 represent only day  conditions on street 3, as shown in Fig. \ref{Scenarios}.
For more convenient description, we will rename above six scenarios as Cases A, B, C, D, E  and F.

The above six scenarios are selected because: 
1) They are all typical V2I mmWave communication scenarios; 
2) the street trend in collected images is relative smooth and gentle, almost parallel to   camera's angle of view; 
3) measurement system   is the same for all cases, including configurations of antennas, cameras, etc..
In addition, they also have certain differences. 
For example,  distance between  base station (i.e. Rx) and  different streets is not exactly the same, which results in different sizes of Tx vehicles in images despite the same viewing angle;
color and size of  Tx vehicles    on three streets are different, 
and more importantly, there will be random non-test vehicles and pedestrians on each street, which may cause non-line-of-sight (NLOS) conditions and affect  radio propagation. 
They are also recorded by RGB camera and further influence  prediction accuracy of model.

The   measurement system consists of  transmitter (Tx) and receiver (Rx) \cite{deepsense}.
Rx includes mmWave receiver, a 60 GHz RF front-end with a 16-element uniform linear array (SIVERS semiconductors), and an RGB camera with 110$^\circ$ field-of-view (FoV) and 30 frames per second (ZED2 from StereoLabs). 
Tx employs a mmWave transmitter with  a 60 GHz quasi-omni antenna (SIVERS semiconductors).
During   measurement process, Rx is fixed, while Tx is installed on a vehicle and moves back and forth along   fixed route in   two-lane streets. During this period, Rx continuously obtains  received power from Tx and captures RGB images. 
Detailed parameters of   measurement system are shown in Table \ref{measure_system}.

\begin{table}[!t]
	\renewcommand{\arraystretch}{1.5}
	\begin{center}
		\caption{Configuration of Measurement System.} 
		\label{measure_system}
		\begin{tabular}{c c}
			\hline
			\hline
			Parameters              & Description \\  
			\hline
			Center Frequency               & 60 GHz \\
			Transmitter antenna     & Omni-directional antenna \\
			Receiver antenna         &  16-Element antenna array \\
			Average sample rate      & 8 Snapshots/s \\
			Resolution of original images & 960$\times$540   \\
			Frame rate of camera     & 30 Frames per second \\
			FoV of camera  & 110$^\circ$ \\
			Average speed of vehicle & 40 km/h  \\
			\hline
			\hline
		\end{tabular}
	\end{center}
\end{table}

\subsection{Data Preprocessing}
Data in   dataset \emph{DeepSense 6G} has been cleaned and pre-processed \cite{deepsense}.
Through methods   such as data synchronization, filtering and verification, it is ensured that data of received power  corresponds to  RGB images, which  meets our experiment requirements, so we basically make no additional change to the  original dataset.
However, it should be pointed out that since our main task is to predict received power through RGB images, in order to simplify calculation amount, although mmWave receiver is a 16-element antenna  array,  we only use  data of one certain channel.
An overview of datasets for training and validation is shown in Fig. \ref{Scenarios}. 

It should be noted that in   datasets we use, there are line-of-sight (LOS) conditions  between  Tx vehicle and  Rx with  90\% probability, and at least  75\% of   time there is only a Tx vehicle exists in  captured images and no other disturbing objects, such as non-test vehicles and pedestrians. 
This is reasonable, because in the wide-angle FoV of  camera, unless it is a highly congested road section, occlusion between vehicles is not obvious, which creates conditions for LOS propagation. 
What's more, whether there are disturbing objects  and whether to identify them in captured images have a huge impact on   prediction accuracy and speed of  model, which will be discussed in detail in Section IV-E.

It also should be explained that since  datasets from \emph{DeepSense 6G} uses only  received power to describe  wireless channel,  prediction of channel characteristics in this work is also limited to  received power.
However,  prediction of received power is still meaningful and important for vehicular communications. 
For example, if  visual data obtained from  base station can be used to help predict  user's received power, it would be able to provide real-time guidance for  base station power control and link budget without taking up additional spectrum, time and space resources, which will greatly improve the safe and efficient operation of vehicular communications.
What's more, 
it is still important to emphasize that this study focuses on cross-field application innovation rather than focusing on  specific channel parameters. 
It is foreseeable that since  received power can be predicted, PL,  delay spread and even Rice K-factor can actually be predicted as long as  dataset is appropriate.

\begin{table}[!t]
	\renewcommand{\arraystretch}{1.5}
	\begin{center}
		\caption{Hyperparameters of proposed two-stage model.} 
		\label{parameter}
		\begin{tabular}{c c}
			\hline
			\hline
			Hyperparameters              & Description \\  
			\hline
			Network for Stage 1    & YOLOv8 \\
			Network for Stage 2  & ResNets \\
			Image size as input   & 224$\times$224 pixels\\
			Number of training epochs               & 50 \\
			Batch size            & 64 \\
			Loss function         &  MSE \\
			Optimization algorithm      & Adam \\
			Activation function & ReLU \\
			Learning rate     & 0.001 \\
			\hline
			\hline
		\end{tabular}
	\end{center}
\end{table}

\subsection{Model Training}
Datasets we use contain   received power and RGB images data of multiple streets as mentioned before, which are divided into training, validation and test sets in different proportions in various experiments. 
Specific dataset division is listed in  Table \ref{exp} and introduced in Section IV.
Original RGB images labeled with  corresponding received power are used to train   two-stage model.
Resolutions of all input images on model training and validation is 960$\times$540$\times$3 pixels (three colour channels, RGB), 
and resolutions of images with BBox, instance segmentation and binary mask  obtained by YOLOv8   are 960$\times$540$\times$3, 960$\times$540$\times$3 and 640$\times$384$\times$1 pixels, respectively, and then they are all resized to  224$\times$224 pixels to match   input size of ResNets.
It should be noted that both YOLOv8 and ResNets have been pre-trained on   COCO dataset and ImageNet2012 dataset by transfer learning, respectively, and  pre-trained weights have been fine-tuned on  training sets.
Through transfer learning (i.e., using a pre-trained existing network architecture),  we can achieve good prediction performance even with a limited dataset.
What's more, it is not to be ignored  that  categories of dynamic scatterers are only vehicles (including Tx vehicle and non-test vehicles) and pedestrians in selected image datasets,  and they are all included in  COCO dataset, thus  pre-trained and fine-tuned YOLOv8 can almost accurately identify  above dynamic scatterers.

Hyperparameters  of YOLOv8 can be referred to \cite{talaat2023improved}. 
And hyperparameters  of prediction model   in stage 2 are shown in  Table \ref{parameter}. 
Number of training epochs  and batch size are 50 and 64 respectively, with learning rate   0.001. 
Adam \cite{kingma2014adam} is selected as optimization algorithm   and  mean square error (MSE) function, which is widely used for regression tasks, is selected as  loss function, with activation function of ReLU.
Both YOLOv8 and ResNets are trained using  deep learning framework PyTorch.

\section{Performance evaluation and analysis}
Based on   \emph{DeepSense 6G}, we carry out a a series of experiments to fully evaluate and analyze the performance of  proposed two-stage model on received power prediction. 
Root mean square error (RMSE) is chosen as   metric to evaluate  prediction accuracy of model.

\begin{table*}
	\renewcommand{\arraystretch}{2.5} 
	\begin{center}
		\caption{Configuration of  five experiments.}
		\label{exp}
		\begin{threeparttable}
			\begin{tabular}{ c| c| c| c| c|  c }
				\hline
				\hline
				Experiments & \makecell[c]{Training and  Validation Set} & Test Set & \makecell[c]{Numbers of \\ Total/Training/Test Data}  &\makecell[c]{Train/Validate/Test \\ Dataset Ratio } & \makecell[c]{Prediction \\ Network}   \\
				\hline
				Scenario Self-Validation    & \multicolumn{2}{c|}{\makecell[c]{Cases A, C, E and F \\ (Day on Streets 1/2/3)}}  
				& 13905 / 12515 / 1390 & 80\% / 10\% / 10\% & ResNet18  \\
				\hline
				Day/Night Cross-Validation  & \makecell[c]{Cases A, C, E and F\\(Day on Streets 1/2/3)}&\makecell[c]{Cases B and D\\(Night on Streets 1/2)} 
				& 18746  / 13905 / 4841 & (Cross Validation)         & ResNet-18  \\
				\hline
				Scenario Cross-Validation & \makecell[c]{Cases A, B, C and D \\ (Day and Night on Streets 1/2)} & \makecell[c]{Case  F \\ (Day on Street 3)} 
				& 14703 / 8739 / 5964 & (Cross Validation)  & ResNet-18 \\
				\hline
				\makecell[c]{Impact of Network Scales} & \makecell[c]{Cases A, C, E and F \\ (Day on Streets 1/2/3)} & \makecell[c]{Case A \\ (Day on Street 1)} 
				& 13905 / 11494 / 2411 & 85\% / 10\% / 5\% & \makecell[c]{ ResNet-18 \\ ResNet-34\\ResNet-50\\ResNet-101\\ResNet-152}  \\
				\hline
				\makecell[c]{ Impact of  Interference \\Elimination} & \multicolumn{2}{c|}{\makecell[c]{Cases C and D \\ (Day and Night on Street 2)}} 
				& 3354 / 3019 / 335 & 80\% / 10\% / 10\% & ResNet-34 \\
				
				\hline
				\hline
				
			\end{tabular}
			
		\end{threeparttable} 
	\end{center}
\end{table*}

\begin{figure*}[!t]
	\centering
	
	\subfloat[]{\includegraphics[width=.33\textwidth]{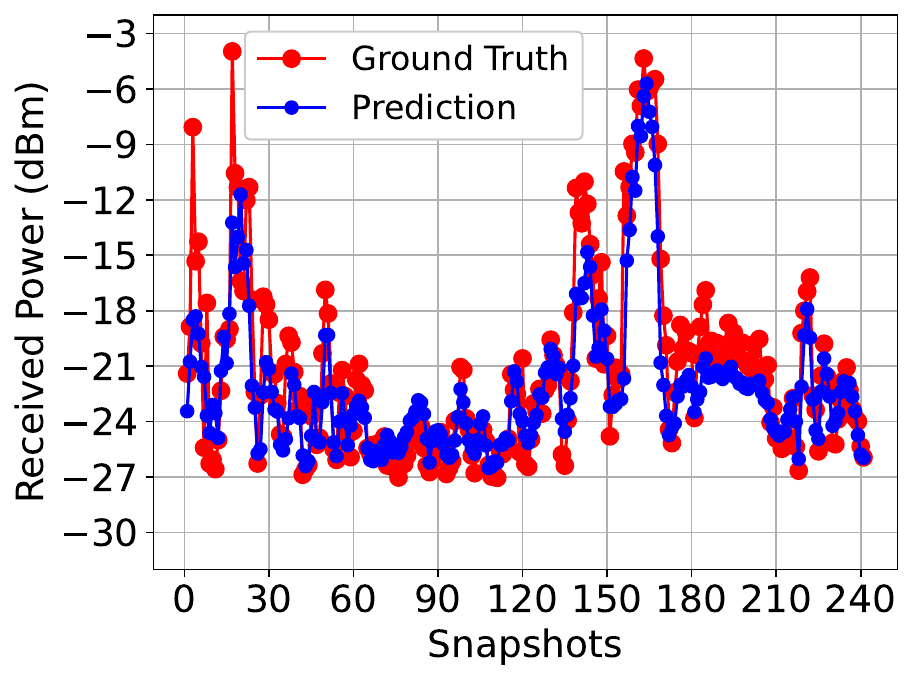}%
		\label{scenario1_dec}}
	\subfloat[]{\includegraphics[width=.33\textwidth]{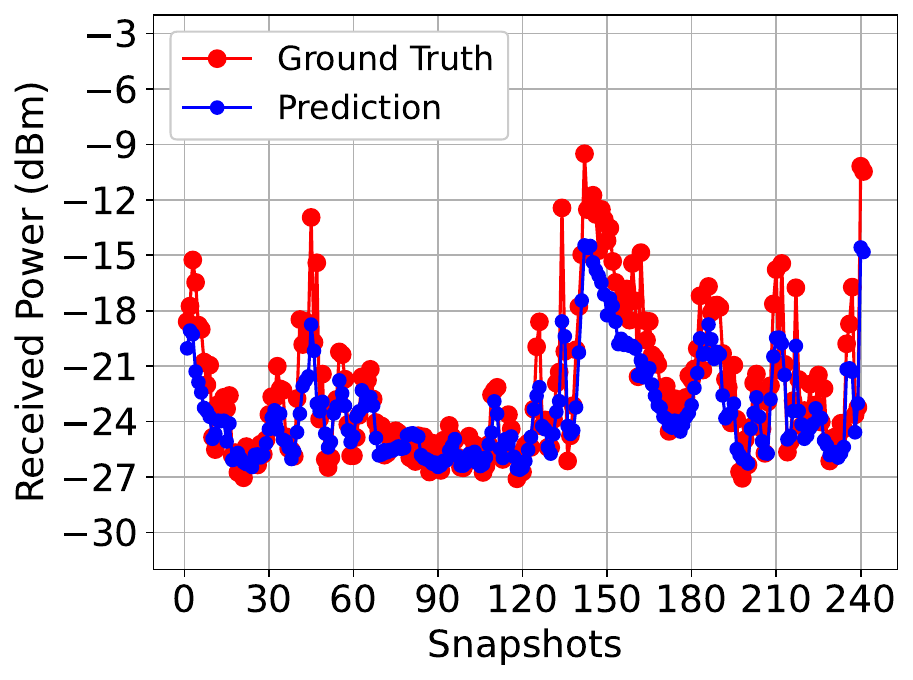}%
		\label{scenario1_seg}}
	\subfloat[]{\includegraphics[width=.33\textwidth]{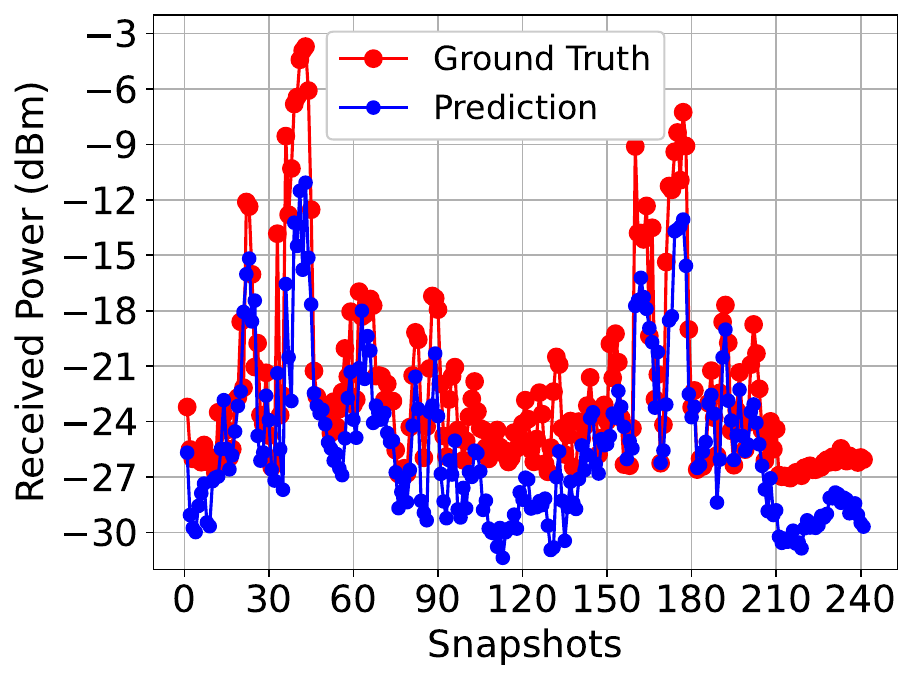}%
		\label{scenario1_mask}}
	\quad
	\caption{
		Prediction results of three kinds of images in Experiment 1.
		Case A of Street 1: (a) BBox;
		(b) Instance Segmentation;
		(c) Binary Mask.
}
	\label{exp1}
\end{figure*}

\begin{figure*}[!t]
	\centering
	
	\subfloat[]{\includegraphics[width=.4\textwidth]{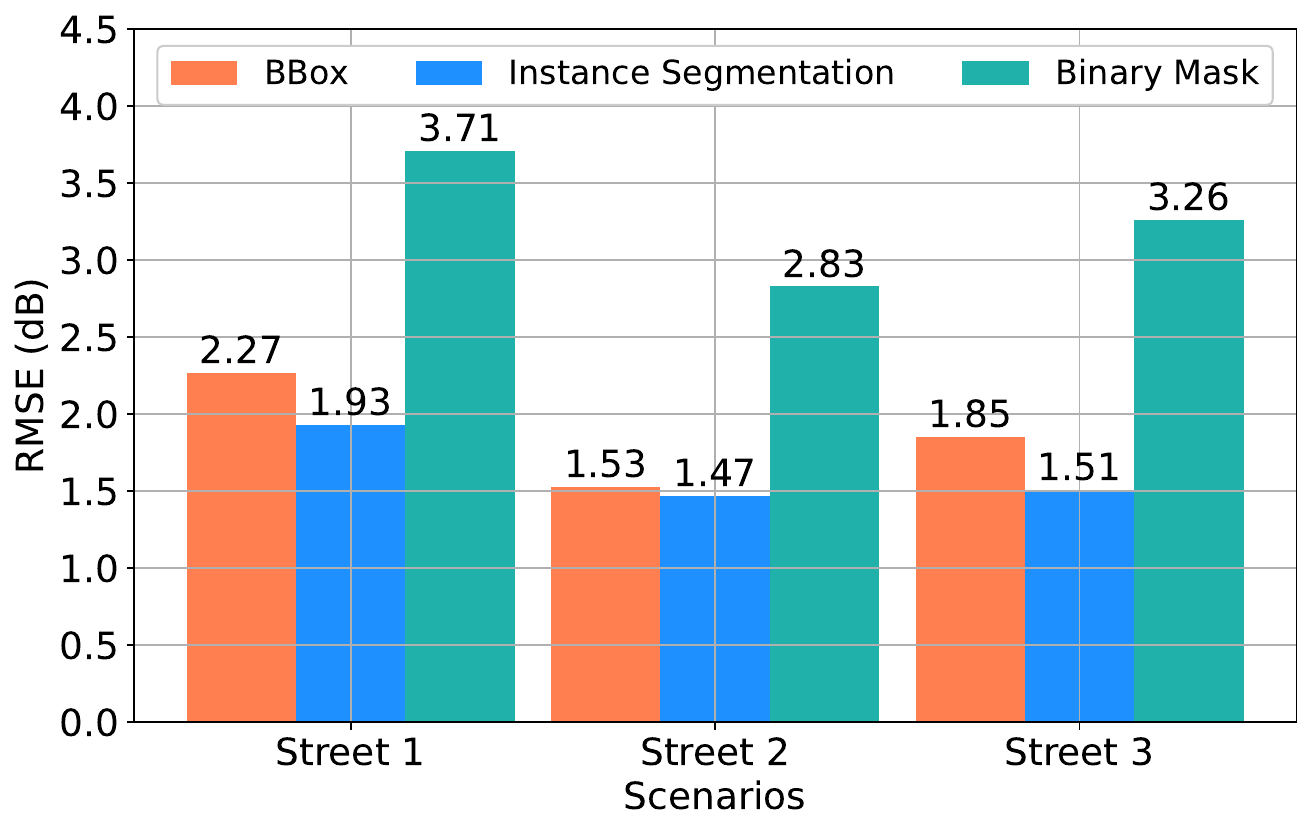}%
		\label{exp1_rmse}}
	\subfloat[]{\includegraphics[width=.4\textwidth]{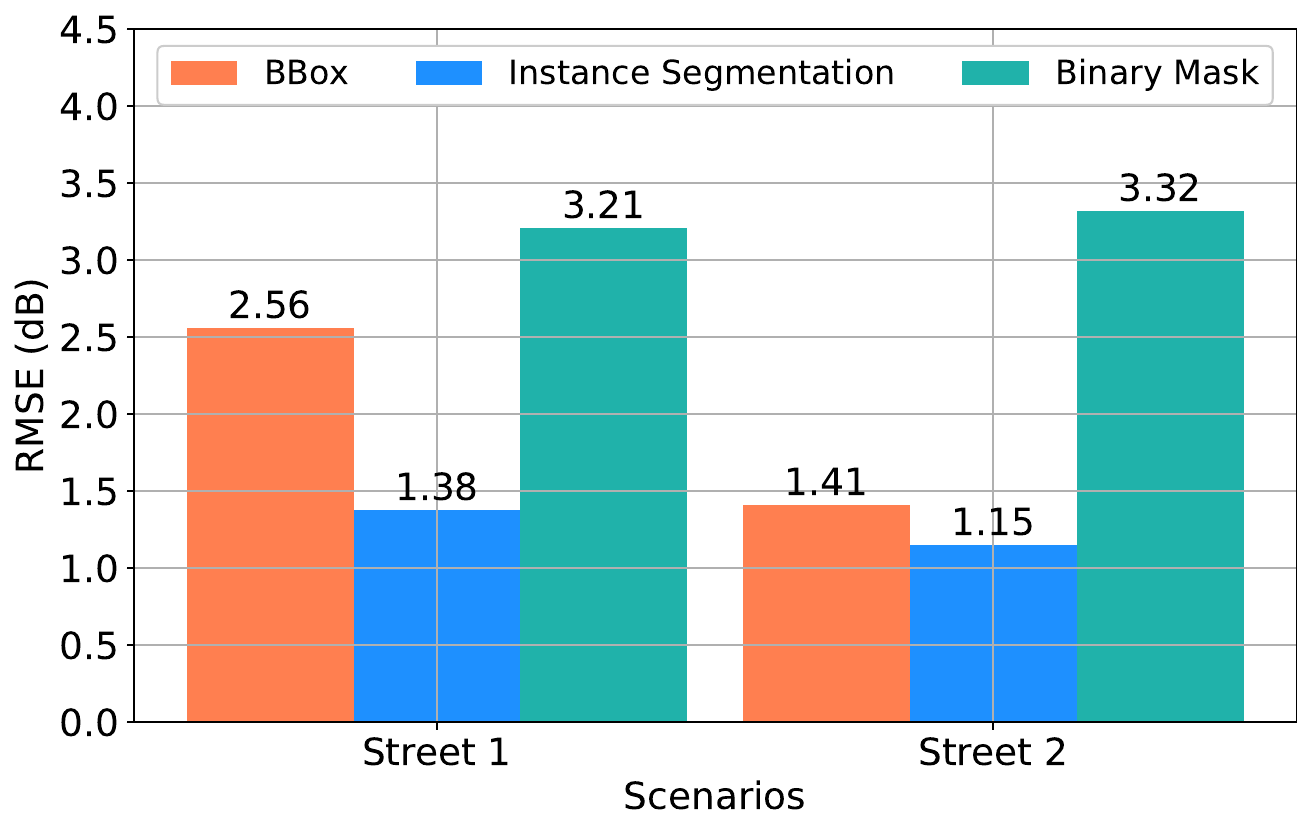}%
		\label{exp2_rmse}}
	\quad
	\caption{
		RMSEs of prediction in (a) : Experiment 1: Scenario Self-Validation   and 
		(b) Experiment 2: Day/Night Cross-Validation.
	}
	\label{RMSEs}
\end{figure*}

\begin{figure*}[!t]
	\centering
	\subfloat[]{\includegraphics[width=.33\textwidth]{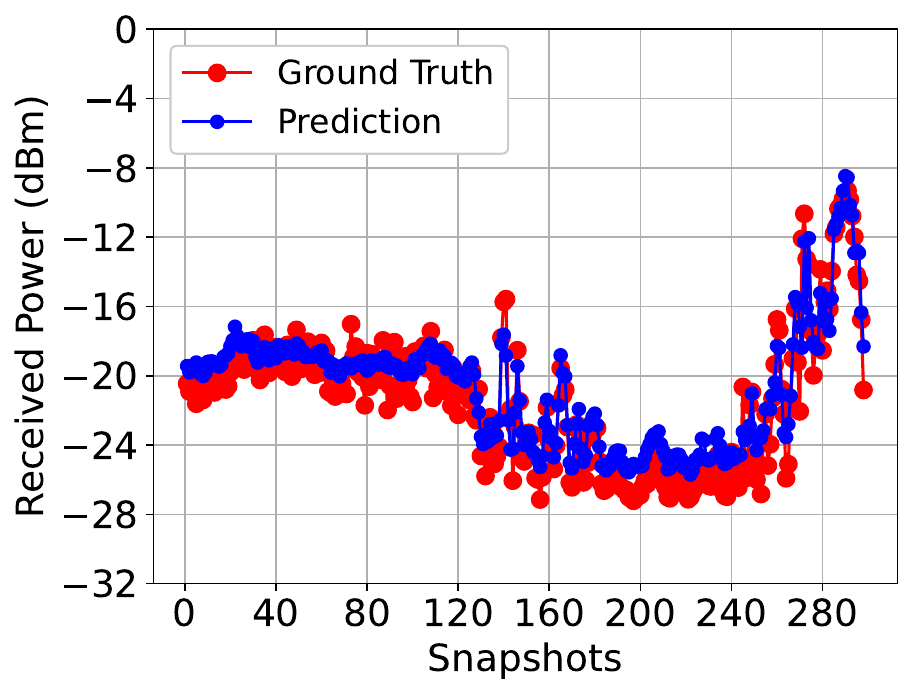}%
		\label{scenario2_dec}}
	\subfloat[]{\includegraphics[width=.33\textwidth]{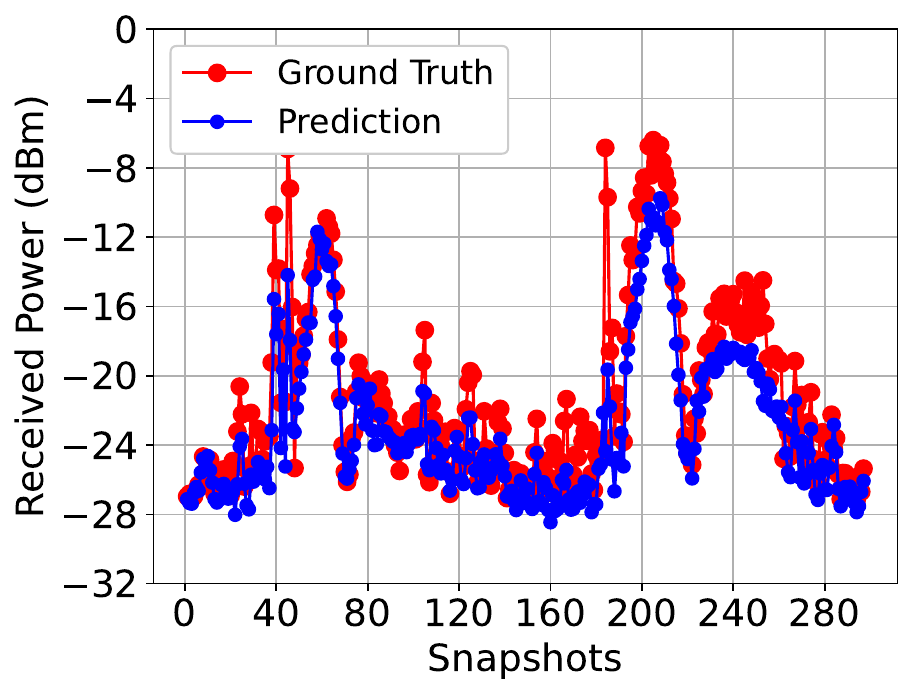}%
		\label{scenario2_seg}}
	\subfloat[]{\includegraphics[width=.33\textwidth]{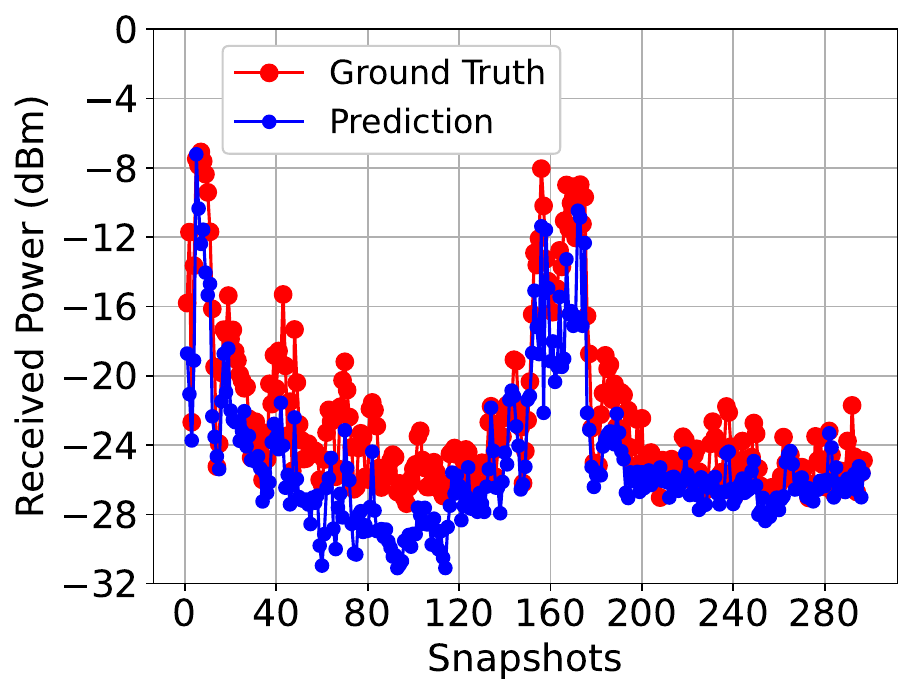}%
		\label{scenario2_mask}}
	\caption{
		Prediction results of three kinds of images in  Experiment 2.
		Case B of Street 1: (a) BBox;
		(b) Instance Segmentation;
		(c) Binary Mask.
	}
	\label{exp2}
\end{figure*}

\subsection{Experiment 1: Scenario Self-Validation }
The purpose of this experiment is to verify  prediction performance of   model on  different test sets after being trained on training sets from the same scenarios as test sets. 
In this experiment, we select four sets of data, cases A, C, E and F, representing  day  conditions of streets 1, 2 and 3,  respectively. 
For  division of datasets,
first we randomly divided data of each case into training set and test set according to the proportion of 90\% and 10\%. 
Then we mix  training sets of  four cases together and randomly divided them into training  and validation sets in proportions  of 80\% and 10\% of total data. 
Finally, We get training, validation and test sets which account for 80\%, 10\% and 10\% of total data respectively, in which test set contains independent data from four cases.

Taking case A  as example,   prediction results are shown in  Fig. \ref{exp1}.
RMSEs of received power prediction  in four cases from three streets when different    images are used  are shown in Fig. \ref{RMSEs}(a). 
It should be noted that three types of images    are used to predict  received power respectively, thus in fact, these are three sub-experiments.   
Division of datasets in each sub-experiment is random, so test sets   of  three sub-experiments are not identical. 
Therefore,  ground truths are not the same in  subfigures in Fig. \ref{exp1}.
For case A, RMSEs of BBox, segmentation and binary mask as inputs to  prediction networks are 2.27, 1.93  and 3.71 dB, respectively.
For streets 2 (Case C) and 3 (Cases E and F), RMSEs are 1.53, 1.47, 2.83 and 1.85, 1.51, 3.26 dB, respectively. 
It can be found that  received power prediction of  three types of images is relatively accurate, and   RMSEs are no more than 4 dB, which indicates that   received power can be predicted  accurately. 
What's more, it is interesting to note that images with instance segmentation has the smallest error, while images with   binary mask has the worst result. 
We suspect that this may be because the former not only highlights  dynamic scatterers, but also retains  static scatterers. 
Therefore,  networks can learn features of environment more fully, while the latter does not contain complete static scatterers and only dynamic scatterers  have been highlighted through white pixels. 
Networks do  not learn   environment  features from images deeply enough, resulting in larger errors.
While images with BBox retain static scatterers, but prominence of dynamic scatterers is slightly inferior to that of instance segmentation, thus   prediction performance is  in the middle.

\subsection{Experiment 2: Day/Night Cross-Validation }

For RGB camera, it can more accurately capture    environmental information during the day due to abundant light, 
while  light at night  will be darker, and  description of   environment in RGB images will be more blurred. 
Therefore, in this cross-validation experiment, we try to use the data   under day conditions to train model, and take   data under night conditions from the same street as   test sets to validate  generalization of proposed model. 
Cases A, C, E and F  from   day   situations of streets 1, 2, 3, are used as  training sets, while   cases B and D  representing day   situations of streets 1 and 2 are selected as test sets.

RMSEs of  prediction  with different  segmentation images are shown in Fig. \ref{RMSEs}(b), and  prediction results on case B are shown in  Fig. \ref{exp2}.
To present  results clearly and visually, only 10\% of data in   case B has been   randomly captured and placed in Fig. \ref{exp2}. 
For street 1  (Case B),  RMSEs of images with BBox, instance segmentation  and binary mask   are 2.56, 1.38, and 3.21 dB, respectively, and for street 2 (Case D),  RMSEs are 1.41, 1.15, 3.32 dB. 
It can be seen that although  prediction network in stage 2 is trained on     datasets of day conditions, when    RGB images at night have been input,  model also can accurately predict   received power.

We can find that similar to  Experiment 1, when images with instance segmentation are used as input,   prediction accuracy is the highest, while images with binary mask have the worst accuracy. 
Reasons for this phenomenon are same as  Experiment 1. 
What's more, an interesting point is that for prediction network in stage 2 trained on the same dataset, no matter whether images in  test sets are from day or night, network can achieve  excellent received power prediction, and their prediction accuracy is not much different. 
On the one hand, it is due to  powerful image processing capability of  prediction network.
On the other hand, this may be due to the fact that in addition to  large differences in pixel colors  between images from day  and night,  both static and dynamic scatterers    in images captured are highly similar.
What's more, artificial light sources such as street lamps and vehicle lights are available at night in urban scenarios,
which reduces the difficulty of  network's recognition of scatterers, 
and  further facilitates prediction network's learning of environment features, 
so that  prediction accuracy of model for day and night images is not subject to wide-scale fluctuations.

\begin{figure*}[!t]
	\centering
	
	\subfloat[]{\includegraphics[width=.33\textwidth]{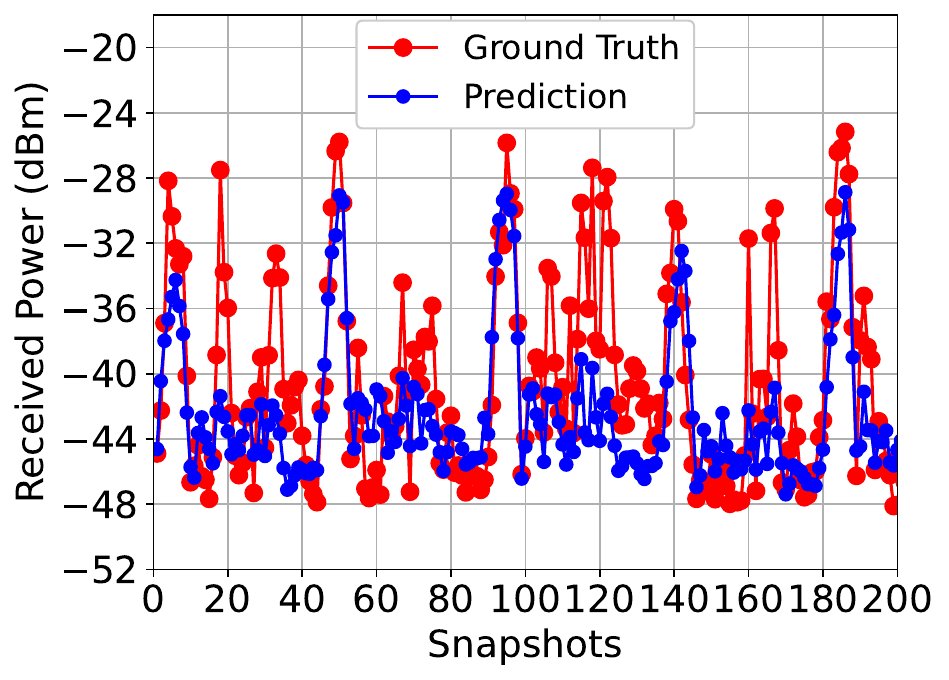}%
		\label{exp3_dec}}
	\subfloat[]{\includegraphics[width=.33\textwidth]{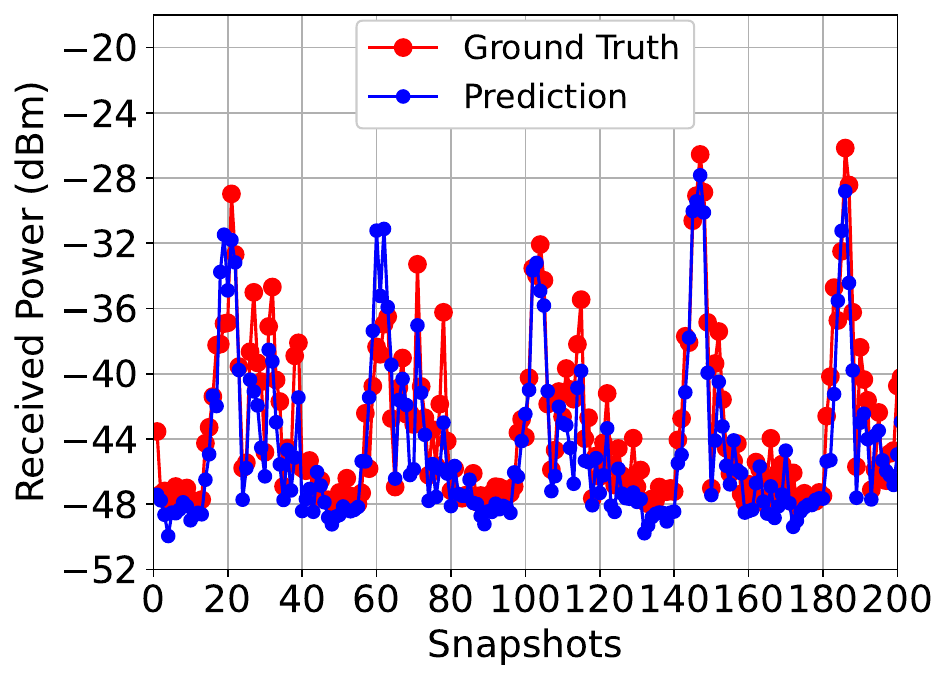}%
		\label{exp3_seg}}
	\subfloat[]{\includegraphics[width=.33\textwidth]{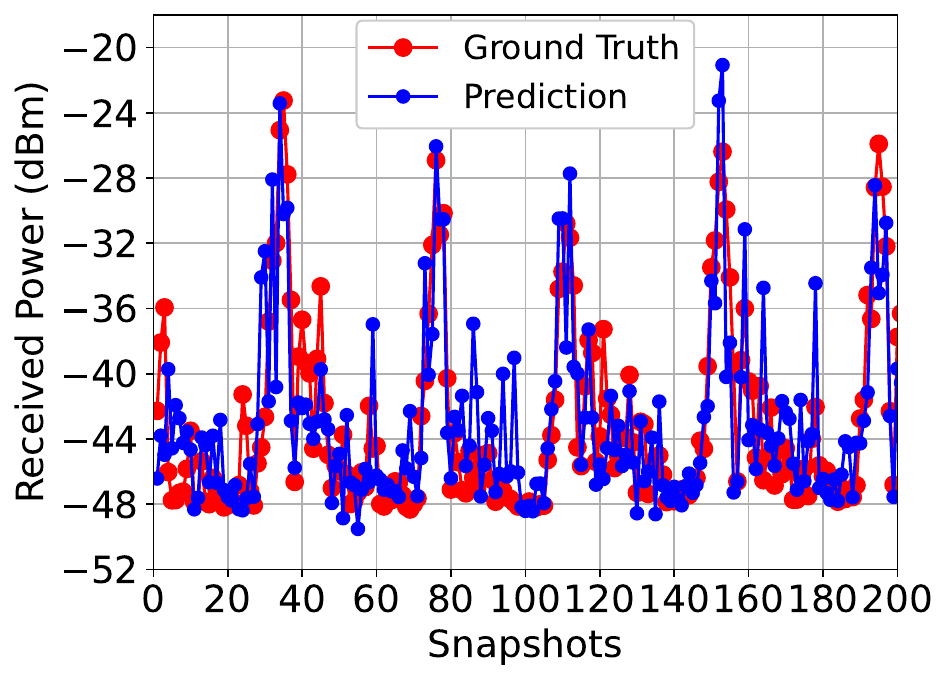}%
		\label{exp3_mask}}
	\caption{
		Prediction results of three kinds of images in Experiment 3.
		(a) BBox;
		(b) Instance Segmentation;
		(c) Binary Mask.
	}
	\label{exp3}
\end{figure*}

\begin{figure*}[!t]
	\centering
	\subfloat[]{\includegraphics[width=.45\textwidth]{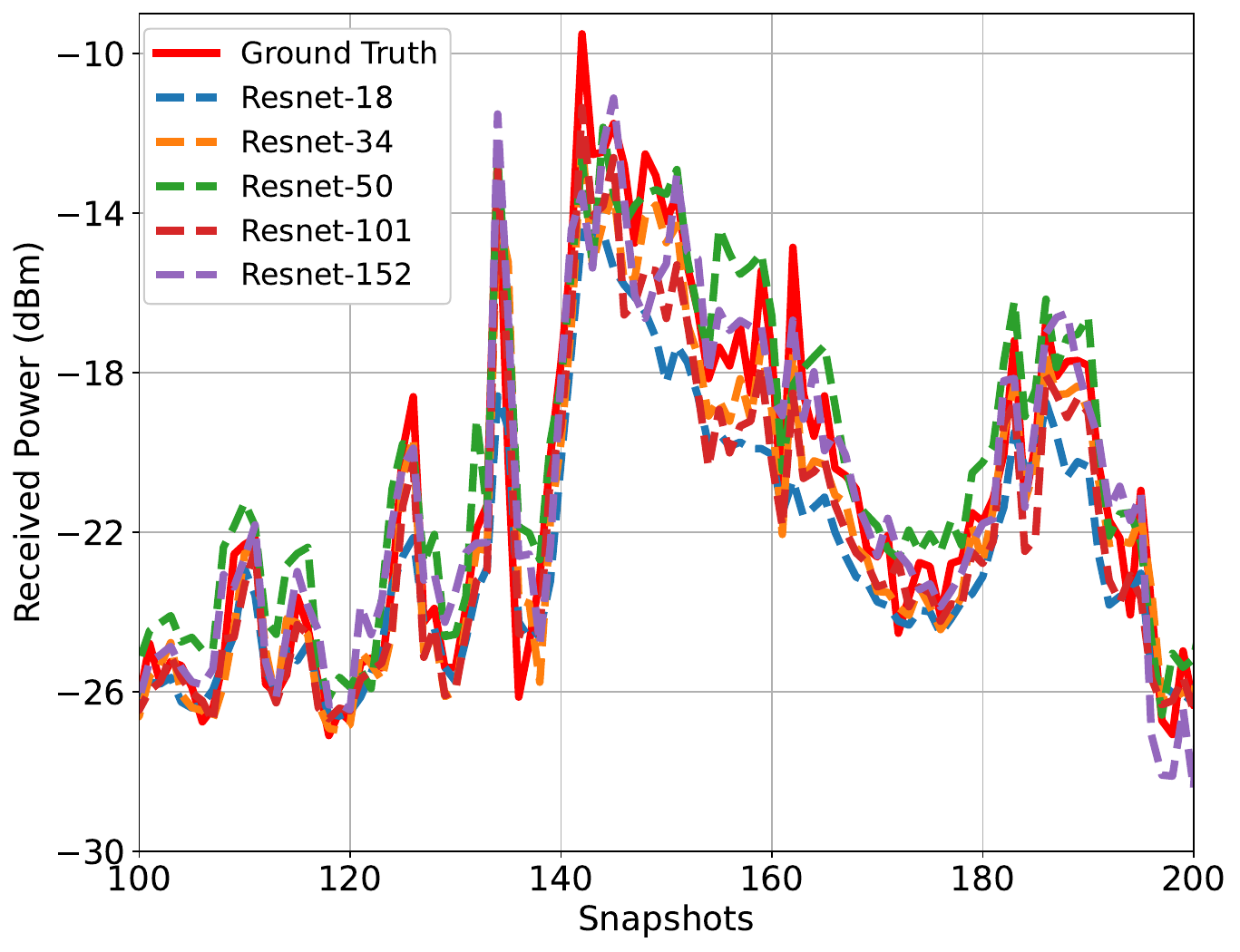}%
		\label{exp4}}
	\subfloat[]{\includegraphics[width=.5\textwidth]{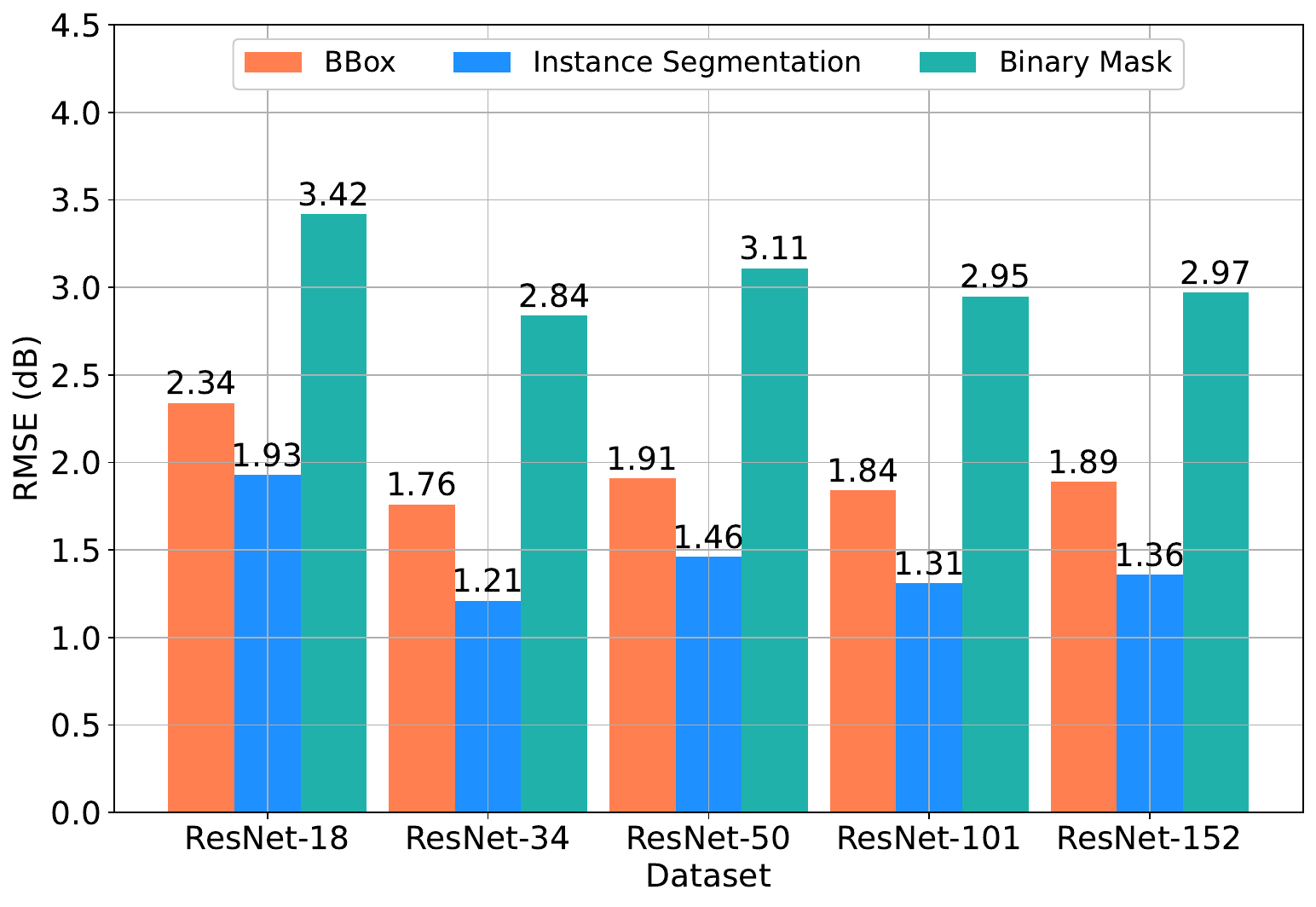}%
		\label{exp4_rmse}}
	\caption{
		Prediction results of images with instance segmentation and RMSEs of different CNNs with three types images in   Experiment 4.
		(a) Prediction results of images with instance segmentation;
		(b) RMSEs of different CNNs with three types images.
	}
	\label{exp44}
\end{figure*}

\subsection{Experiment 3: Scenario Cross-Validation }
The purpose  is to verify generalization ability of trained model on test sets from new scenarios in this experiment. 
Cases A, B, C, D are selected as  training set from  day and night of streets 1 and 2, respectively, and cases E, F from street 3 during the day are selected as  test set. 

Similarly, we still use  three types of   images  to evaluate   performance.
Prediction results of BBox,  instance segmentation and binary mask have been shown in Fig. \ref{exp3} and their RMSEs are 4.46, 3.42 and 3.76 dB, respectively, only 200 snapshots are included in the figure for ease of observation. 
It can be seen that three networks can still roughly predict the trend of  received power, but compared with Experiments 1 and 2,  prediction accuracy is somewhat decreased. 
Prediction error on images with instance segmentation is still the smallest, while  images with BBox is the largest, and their RMSEs are increased by 2-3 dB. 
However,  RMSE of images with binary mask basically does not change, which indicates that  images from new scenarios has little impact on prediction network trained on images with binary mask. 
This is due to characteristics of binary mask itself, in which only dynamic scatterers are additionally displayed with white areas, while static environment is covered by black areas, thus  scenario changes are not actually visible in   images with binary mask. 
What's more, it should be noted that  decline in prediction performance of  above three networks may also come from differences between  data sets. 
In  Figs. \ref{exp1} and \ref{exp2},  values of  received power fluctuates between -30 dB and 0 dB, while in Fig. \ref{exp3}, the value varies from -50 dB to -20 dB, and prediction networks are trained on   data of the former and tested on   the latter. 
The inconsistencies   between  received power ranges in   training and test sets lead to  degradation of    prediction performance.

\subsection{Experiment 4: Impact of Network Scales }
In fact,    network in stage 1 is only responsible for converting  original RGB images into images with environment, which does not directly affect prediction accuracy of model, because YOLOv8 after pre-trained  and fine-tuned      can almost achieve all of image conversion, 
and network in stage 2 really affects prediction accuracy of the whole model. 
In the first three experiments, we have used    Resnet-18,
which achieves relatively accurate prediction results. 
In this experiment, we try to use different layers of ResNets as prediction networks to validate  impact of   network scales on model performance. 
Networks we choose include ResNet-18, ResNet-34, ResNet-50, ResNet-101  and ResNet-152, with network depths of 18, 34, 50, 101  and 152 layers, respectively.
Datasets of cases A, C, E and F are still selected, and divided methods of datasets are similar to Experiment 1, except that test set this time only comes from case A,   day  condition on street 1. 
We randomly selected 30\% of  data in case A as  test set, with a total of 723 sets of data, accounting for about 5\% of total data volume. 
The other training  and validation sets are  based on total data volume, divided into 85\% and 10\%.

Results of  received power prediction   using images with instance segmentation as an example and RMSEs of different CNNs with three types images in this experiment are shown in Fig. \ref{exp44}.
To present  results clearly, only 100 snapshots in   test set are shown in Fig. \ref{exp44}(a). 
As can be seen from Fig. \ref{exp44}(a), ResNets with different scales have better predictive effects on   received power and can predict   trend of  received power. 
Even so, it also  can be found   that RMSEs of different CNNs  still have gaps. 
Among them, ResNet-34 has the smallest RMSEs,
the second  is ResNet-50, 
followed by ResNet-151 and 152, they are very close, 
while the largest RMSEs are from ResNet-18.
It can be seen that in the case of three types of images being input to predict  received power, with the deepening of CNNs,   prediction performance will be improved, but when a certain threshold is reached, if   network scale continues to deepen, it will not only bring additional computing costs, but also obtain fairly small benefits. 
For    ResNet-151 and ResNet-152, their networks differ by 51 layers, but   RMSE only increases by 0.02-0.05 dB. 
However, decreasing   network scale will also lead to poorer performance, thus reasonable trade-off is necessary.

\begin{figure}[!t]
	\centering
	\includegraphics[width=.48\textwidth]{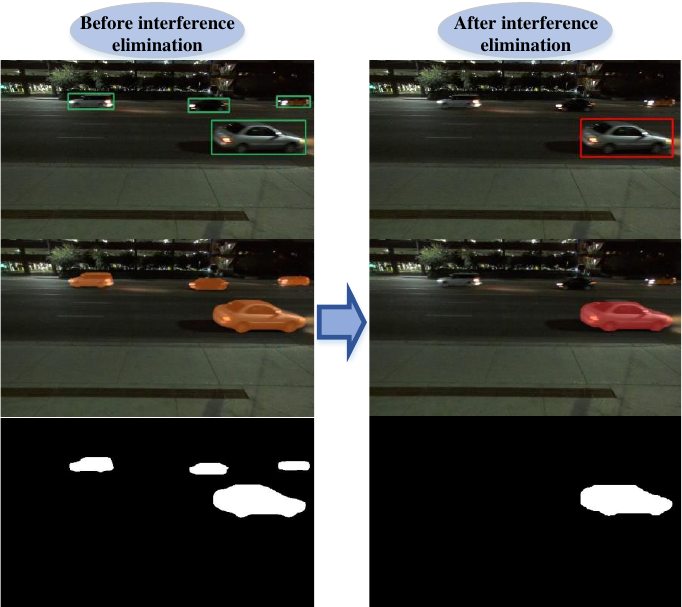}%
	\caption{Image comparison before and after interference elimination.}                    
	\label{remove}
\end{figure}

\begin{figure}[!t]
	\centering
	\includegraphics[width=.45\textwidth]{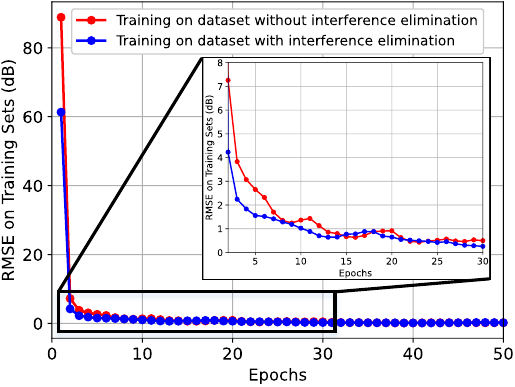}%
	\caption{RMSEs on different training sets in Experiment 5.}                    
	\label{exp5_loss}
\end{figure}

\subsection{Experiment 5: Impact of Interference Elimination}
Three types of image  highlight features of each dynamic scatterer in previous four experiments,  including Tx vehicle, non-test vehicles and pedestrians, 
while YOLOv8  do not accurately identify  Tx vehicle and only highlight it. 
In this experiment, we generate images in which only  Tx vehicle is highlighted, while other non-test dynamic scatterers are no longer highlighted like static  scatterers, which called interference elimination.
Then we use these images to train prediction network to validate prediction performance of  model after interference elimination.

To achieve above purpose, we need to retrain YOLOv8   and we have to manually label  original RGB images. 
Datasets we selected are cases C and D, representing both day and night conditions of street 2, with a total of 3522 data. 
First, we use   tool    Labelme \cite{labelme}  to draw rectangular boxes around  Tx vehicle in order to generate BBox, and then using software  X-Anylabeling  \cite{X-AnyLabeling}  to outline Tx vehicle in order to generate segmentation and binary mask. 
In fact, we only manually label about 2000 images and use them to train  YOLOv8, and remaining approximately 1500  images are automatically generated by YOLOv8 after  training. 
Through above steps, we obtain   images of only Tx vehicle highlighted, as shown in Fig. \ref{remove}.
We have randomly divided them into training, validation and test sets according to ratio of 80\%, 10\% and 10\%, 
without distinguishing   day and night conditions.
In stage 2, we choose Resnet-34 as  prediction network because it has been proved to have the best performance in Experiment 4.

\begin{figure*}[!t]
	\centering
	\subfloat[]{\includegraphics[width=.48\textwidth]{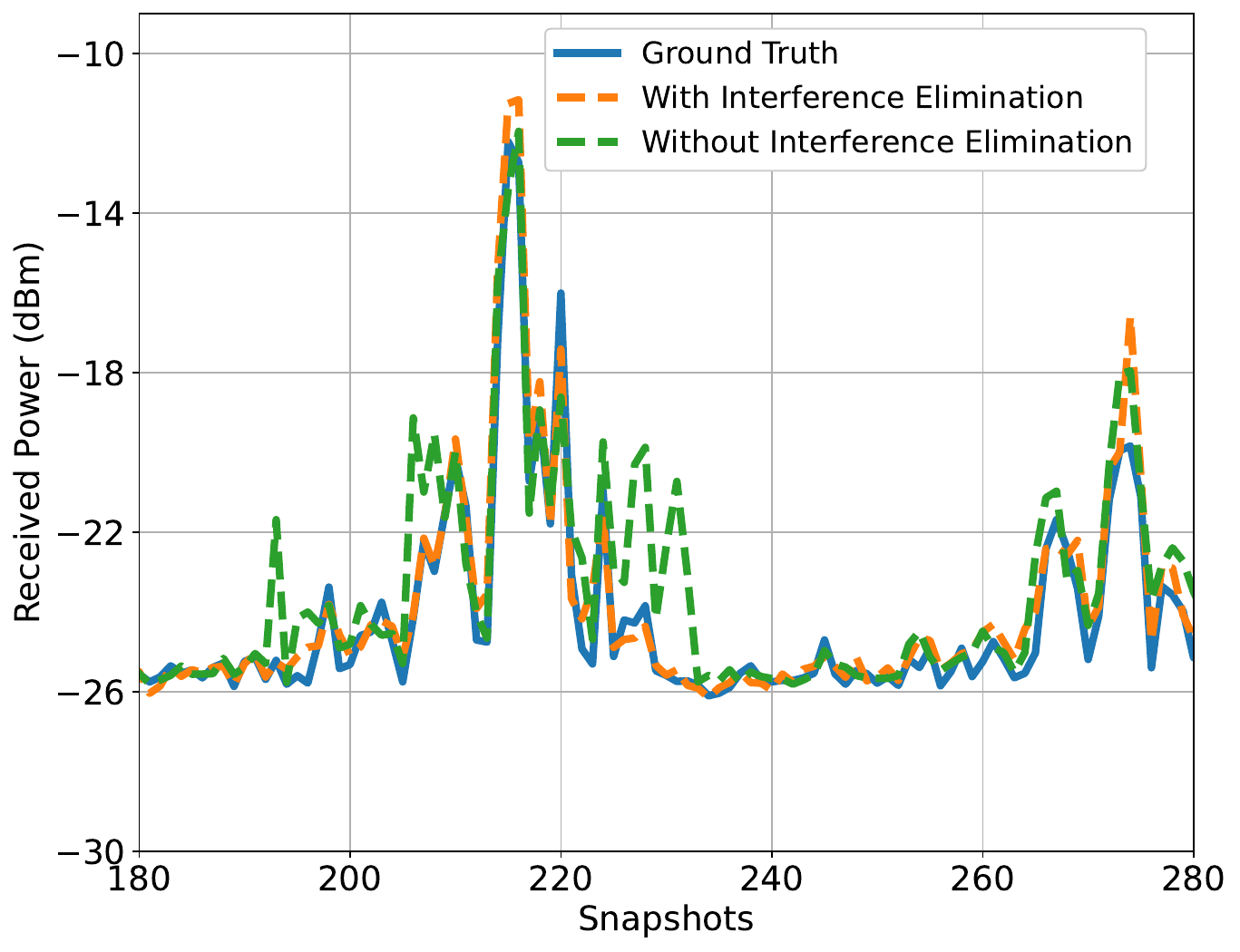}%
		\label{exp5}}
	\subfloat[]{\includegraphics[width=.46\textwidth]{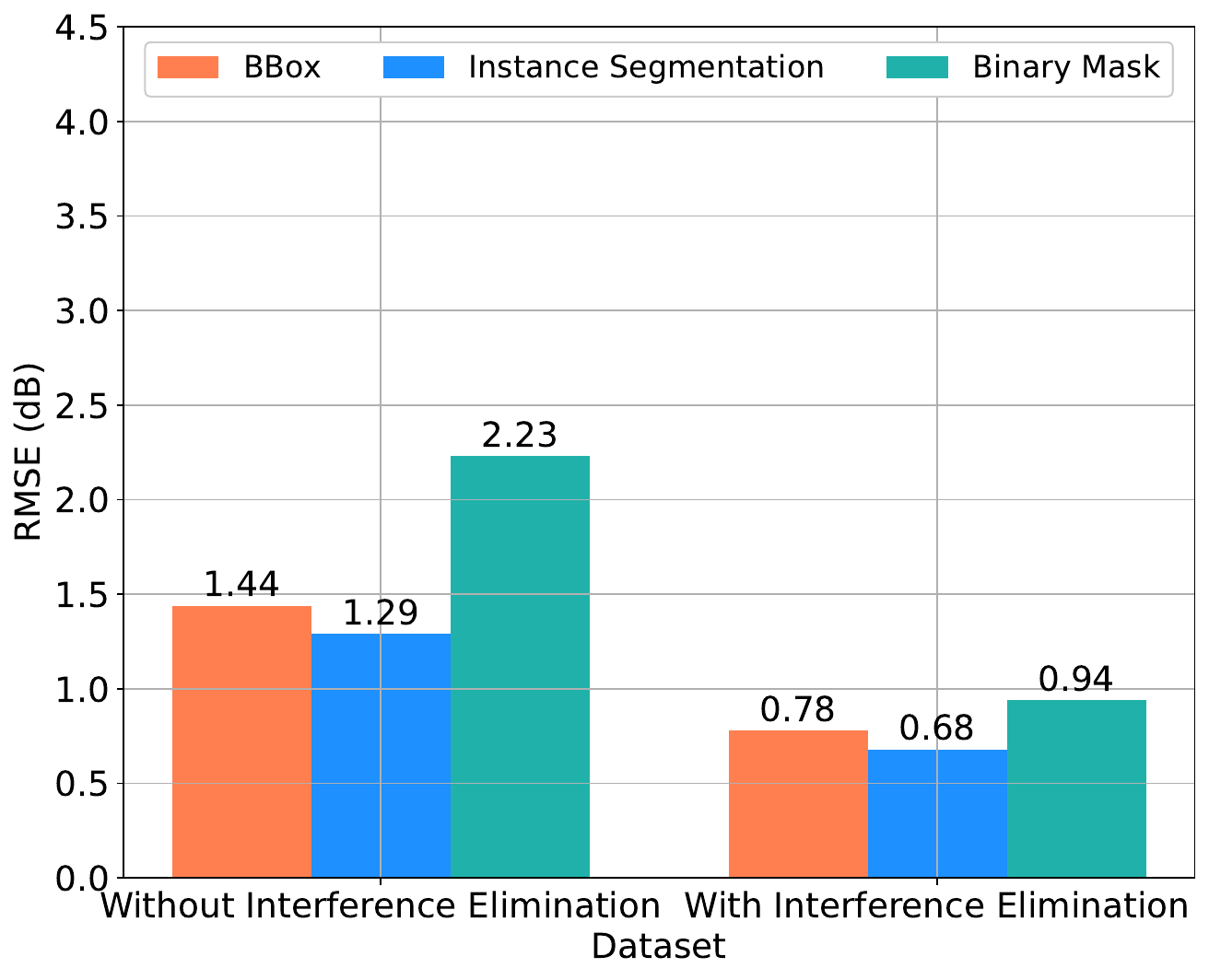}%
		\label{exp5_rmse}}
	\caption{
		Prediction results of images with instance segmentation and RMSEs of before/after interference elimination with three types images in  Experiment 5.
		(a) Prediction results of images with instance segmentation;
		(b) RMSEs of before/after interference elimination with three types images.
	}
	\label{exp55}
\end{figure*}

Results of  received prediction   using images with instance segmentation as an example and RMSEs of two conditions with three types images in this experiment are shown in Figs. \ref{exp55}(a) and (b).
Only 100 snapshots in   test set are shown in Fig. \ref{exp55}(a) for ease of observation.
It can be seen that  prediction network trained using   images after interference elimination has better performance.   RMSEs   on  images with BBox, instance segmentation  and binary mask are 0.78, 0.68  and 0.94 dB, respectively, and RMSEs of with  images before interference elimination are 1.45, 1.19 and 2.13 dB, respectively.
Fig. \ref{exp5_loss} demonstrates  RMSEs of each epoch when   model is trained on    datasets with and without interference elimination. 
It can be found that  model trained on both datasets converges at epoch 30 and interference elimination is more conducive to improving  convergence speed and prediction accuracy of  model.
We can find that   model has more powerful prediction performance when  features of exclusive target object is labeled, while too many non-target objects are labeled, which brings  burden to  prediction network, resulting in a decline in  prediction speed and accuracy of   network. 

\subsection{Discussion}
To adequately evaluate   performance of   model, we have conducted five experiments to verify it from different aspects.
First, we verify  prediction accuracy  on  test sets from different scenarios in Experiment 1, with RMSE arrange from 1.47 to 3.71 dB. 
Secondly, cross-validation has been used to verify   generalization in Experiments 2 and 3.
Networks trained using day data from the same street work well for night predictions, with RMSE between 1.15-3.32 dB.
Although accuracy of prediction networks tested on    new scenarios will decrease,  trend of  received power can still  be predicted   generally   with RMSEs between 3.42-4.46 dB.
Results of these two experiments prove that the proposed model has fairly good generalization.
Then, effects of   network scale in stage 2  on prediction accuracy have been discussed    in Experiment 4. 
The results demonstrate that  relationship between  network scale  and  prediction accuracy is not linear, and reasonable tradeoffs are needed.
Moreover, with moderate number of network layers and best prediction accuracy, Resnet-34 should be prioritized   for channel prediction tasks.
Finally, we analyze  impacts of interference elimination on  prediction results in  stage 1 in Experiment 5. 
The results show that eliminating interference and keeping only features of target object will be more beneficial to  prediction accuracy and efficiency of the model.
Therefore, for   V2I  scenarios of  single user, accurately identifying  features of   user will be more conducive to   predict wireless channel.
Moreover, it should not be ignored that in all five experiments,  highest prediction accuracy is achieved when images with instance segmentation are input in stage 2, while  images with  binary mask have the worst accuracy. 
Therefore,  instance segmentation-based method   will be more beneficial to received power prediction.

\section{CONCLUSION}
In this paper, vision-aided two-stage model based on CV for  channel prediction in  mmWave vehicular communication scenarios has been proposed.
Firstly,  original images about different propagation environment are obtained through  RGB camera,
 and three methods of environment information extraction based on CV are applied.
Secondly, a vision-aided two-stage channel prediction model has been established. 
Finally, five experiments are carried out based on measurement datasets to evaluate  performance of the proposed model.
Experiment  results have validated the excellent performance   in practicability, reliability  and generalization.
The model in this paper has the advantages of high accuracy, simple operation  and high implementation efficiency.
It is achieved by using only RGB images of propagation environment to predict  received power  of  user.
The ideas in this paper can be extended to   prediction of other typical channel characteristics.  It will be  helpful to promote   further deep integration of AI (especially CV) technology and wireless communication and also provide a set of novel solutions for     realization of intelligent  vehicular communication system.
   
   \balance
   \bibliographystyle{IEEEtran}

   \nocite{*}
   
   \bibliography{IEEEabrv,ref}

\begin{thebibliography}{10}
\providecommand{\url}[1]{#1}
\csname url@samestyle\endcsname
\providecommand{\newblock}{\relax}
\providecommand{\bibinfo}[2]{#2}
\providecommand{\BIBentrySTDinterwordspacing}{\spaceskip=0pt\relax}
\providecommand{\BIBentryALTinterwordstretchfactor}{4}
\providecommand{\BIBentryALTinterwordspacing}{\spaceskip=\fontdimen2\font plus
\BIBentryALTinterwordstretchfactor\fontdimen3\font minus
  \fontdimen4\font\relax}
\providecommand{\BIBforeignlanguage}[2]{{%
\expandafter\ifx\csname l@#1\endcsname\relax
\typeout{** WARNING: IEEEtran.bst: No hyphenation pattern has been}%
\typeout{** loaded for the language `#1'. Using the pattern for}%
\typeout{** the default language instead.}%
\else
\language=\csname l@#1\endcsname
\fi
#2}}
\providecommand{\BIBdecl}{\relax}
\BIBdecl

\bibitem{APS2024}
X.~Zhang, R.~He, M.~Yang, Z.~Zhang, Z.~Qi, and B.~Ai, ``Vision-aided channel
  prediction with {RGB} images for mmwave communications,'' in \emph{2024 IEEE
  International Symposium on Antennas and Propagation and INC/USNC‐URSI Radio
  Science Meeting (AP-S/INC-USNC-URSI)}, 2024, pp. 1845--1846.

\bibitem{itu1}
\BIBentryALTinterwordspacing
{ITU-R\vspace{0mm}}, ``{Future technology trends of terrestrial International
  Mobile Telecommunications systems towards 2030 and beyond},'' \emph{{Report
  M.2516-0}}, Nov. 2022. [Online]. Available:
  \url{https://www.itu.int/dms_pub/itu-r/opb/rep/R-REP-M.2516-2022-PDF-E.pdf}
\BIBentrySTDinterwordspacing

\bibitem{itu2}
{\vspace{0mm}ITU-R}, ``{Framework and overall objectives of the future
  development of IMT for 2030 and beyond},'' \emph{{DRAFT NEW RECOMMENDATION}},
  Jun. 2023.

\bibitem{Huang2022}
C.~Huang \emph{et~al.}, ``Artificial intelligence enabled radio propagation for
  communications—{Part I}: Channel characterization and antenna-channel
  optimization,'' \emph{{IEEE} Trans. Antennas Propag.}, vol.~70, no.~6, pp.
  3939--3954, 2022.

\bibitem{hrs2025}
R.~He, N.~D. Cicco, B.~Ai, M.~Yang, Y.~Miao, and M.~Boban, ``{COST CA20120
  INTERACT} framework of artificial intelligence-based channel modeling,''
  \emph{IEEE Wirel. Commun.}, pp. 1--8, 2025.

\bibitem{Graff2023}
A.~Graff, Y.~Chen, N.~González-Prelcic, and T.~Shimizu, ``Deep learning-based
  link configuration for radar-aided multiuser mmwave vehicle-to-infrastructure
  communication,'' \emph{{IEEE} Trans. Veh. Technol.}, vol.~72, no.~6, pp.
  7454--7468, 2023.

\bibitem{Yuxin2023}
Y.~Zhang \emph{et~al.}, ``Generative adversarial networks based digital twin
  channel modeling for intelligent communication networks,'' \emph{China
  Commun.}, vol.~20, no.~8, pp. 32--43, 2023.

\bibitem{huangchen}
C.~\vspace{0mm}Huang \emph{et~al.}, ``Artificial intelligence enabled radio
  propagation for communications—part {II}: Scenario identification and
  channel modeling,'' \emph{{IEEE} Trans. Antennas Propag.}, vol.~70, no.~6,
  pp. 3955--3969, 2022.

\bibitem{Qiu2024}
Z.~Qiu, R.~He, M.~Yang, S.~Zhou, L.~Yu, C.~Wang, Y.~Zhang, J.~Fan, and B.~Ai,
  ``{CNN}-based path loss prediction with enhanced satellite images,''
  \emph{{IEEE} Antennas Wireless Propag. Lett.}, vol.~23, no.~1, pp. 189--193,
  2024.

\bibitem{yangmi}
M.~Yang, R.~He, B.~Ai, C.~Huang, C.~Wang, Y.~Zhang, and Z.~Zhong,
  ``{AI}-enabled data-driven channel modeling for future communications,''
  \emph{{IEEE} Commun. Mag.}, pp. 1--7, 2023.

\bibitem{QI2023241}
Z.~Qi \emph{et~al.}, ``Point cloud-based environment reconstruction and ray
  tracing simulations for railway tunnel channels,'' \emph{High-speed Railway},
  vol.~1, no.~4, pp. 241--247, 2023.

\bibitem{Yang2021}
M.~Yang \emph{et~al.}, ``Machine-learning-based scenario identification using
  channel characteristics in intelligent vehicular communications,''
  \emph{{IEEE} Trans. Intell. Transp. Syst.}, vol.~22, no.~7, pp. 3961--3974,
  2021.

\bibitem{tian}
Y.~Tian, G.~Pan, and M.-S. Alouini, ``Applying deep-learning-based computer
  vision to wireless communications: Methodologies, opportunities, and
  challenges,'' \emph{IEEE Open J. Commun. Society}, vol.~2, pp. 132--143,
  2021.

\bibitem{Nishio2021}
T.~Nishio, Y.~Koda, J.~Park, M.~Bennis, and K.~Doppler, ``When wireless
  communications meet computer vision in beyond {5G},'' \emph{IEEE
  Communications Standards Magazine}, vol.~5, no.~2, pp. 76--83, 2021.

\bibitem{he2019applications}
R.~He and Z.~Ding, \emph{Applications of machine learning in wireless
  communications}, 2019, vol.~81.

\bibitem{MiYang2023}
M.~Yang \emph{et~al.}, ``Dynamic {V2V} channel measurement and modeling at
  street intersection scenarios,'' \emph{IEEE Trans. Antennas Propagat.},
  vol.~71, no.~5, pp. 4417--4432, May 2023.

\bibitem{hrs2020}
R.~He \emph{et~al.}, ``Propagation channels of {5G} millimeter-wave
  vehicle-to-vehicle communications: {Recent} advances and future challenges,''
  \emph{IEEE Veh. Technol. Mag.}, vol.~15, no.~1, pp. 16--26, 2020.

\bibitem{cje}
Y.~Jin and D.~Wang, ``Study on static deflection model of {MEMS} capacitive
  microwave power sensors,'' \emph{Chinese J Electron}, vol.~33, no.~5, pp.
  1188--1195, 2024.

\bibitem{Zhang2024}
X.~Zhang, R.~He, M.~Yang, Z.~Qi, Z.~Zhang, B.~Ai, and R.~Chen, ``Narrowband
  channel measurements and statistical characterization in subway tunnels at
  1.8 and 5.8 {GHz},'' \emph{{IEEE} Trans. Veh. Technol.}, pp. 1--13, 2024.

\bibitem{Zhang2023}
X.~Zhang \emph{et~al.}, ``Measurements and modeling of large-scale channel
  characteristics in subway tunnels at 1.8 and 5.8 {GHz},'' \emph{{IEEE}
  Antennas Wireless Propag. Lett.}, vol.~22, no.~3, pp. 561--565, 2023.

\bibitem{He2013}
R.~He, Z.~Zhong, B.~Ai, J.~Ding, Y.~Yang, and A.~F. Molisch, ``Short-term
  fading behavior in high-speed railway cutting scenario: Measurements,
  analysis, and statistical models,'' \emph{{IEEE} Trans. Antennas Propag.},
  vol.~61, no.~4, pp. 2209--2222, 2013.

\bibitem{Zhengyu2023}
Z.~Zhang, R.~He, B.~Ai, M.~Yang, X.~Zhang, R.~Chen, H.~Zhang, and Z.~Zhong, ``A
  shared multipath components evolution model for integrated sensing and
  communication channels,'' \emph{{IEEE} Antennas Wireless Propag. Lett.},
  vol.~22, no.~12, pp. 2975--2978, 2023.

\bibitem{YiZeng2020}
Y.~Zeng \emph{et~al.}, ``Measurement and simulation for
  vehicle-to-infrastructure communications at 3.5 {{GHz}} for {{5G}},''
  \emph{Wirel. Commun. Mob. Comput.}, vol. 2020, pp. 1--13, Dec. 2020.

\bibitem{MiYang2020}
M.~Yang \emph{et~al.}, ``Measurements and cluster-based modeling of
  vehicle-to-vehicle channels with large vehicle obstructions,'' \emph{IEEE
  Trans. Wireless Commun.}, vol.~19, no.~9, pp. 5860--5874, Sep. 2020.

\bibitem{Wang2018}
Y.~Wang, K.~Venugopal, A.~F. Molisch, and R.~W. Heath, ``{MmWave}
  vehicle-to-infrastructure communication: {Analysis} of urban microcellular
  networks,'' \emph{{IEEE} Trans. Veh. Technol.}, vol.~67, no.~8, pp.
  7086--7100, 2018.

\bibitem{Longhe2019}
L.~Wang, B.~Ai, D.~He, K.~Guan, J.~Zhang, J.~Kim, and Z.~Zhong,
  ``Vehicle-to-infrastructure channel characterization in urban environment at
  28 {GHz},'' \emph{China Commun.}, vol.~16, no.~2, pp. 36--48, 2019.

\bibitem{Wang2022}
C.-X. Wang, Z.~Lv, X.~Gao, X.~You, Y.~Hao, and H.~Haas, ``Pervasive wireless
  channel modeling theory and applications to {6G} {GBSMs} for all frequency
  bands and all scenarios,'' \emph{{IEEE} Trans. Veh. Technol.}, vol.~71,
  no.~9, pp. 9159--9173, 2022.

\bibitem{Wang2023}
C.-X. Wang, Z.~Lv, Y.~Chen, and H.~Haas, ``A complete study of
  space-time-frequency statistical properties of the {6G} pervasive channel
  model,'' \emph{{IEEE} Trans. Commun.}, vol.~71, no.~12, pp. 7273--7287, 2023.

\bibitem{alrabeiah2020viwi}
M.~Alrabeiah, J.~Booth, A.~Hredzak, and A.~Alkhateeb, ``Viwi vision-aided
  mmwave beam tracking: Dataset, task, and baseline solutions,'' \emph{arXiv
  preprint arXiv:2002.02445}, 2020.

\bibitem{Alrabeiah2020}
M.~Alrabeiah, A.~Hredzak, and A.~Alkhateeb, ``Millimeter wave base stations
  with cameras: Vision-aided beam and blockage prediction,'' in \emph{Proc.
  IEEE 91st Veh. Technol. Conf. (VTC-Spring)}, 2020, pp. 1--5.

\bibitem{Charan2021}
G.~Charan, M.~Alrabeiah, and A.~Alkhateeb, ``Vision-aided {6G} wireless
  communications: Blockage prediction and proactive handoff,'' \emph{{IEEE}
  Trans. Veh. Technol.}, vol.~70, no.~10, pp. 10\,193--10\,208, 2021.

\bibitem{Nishio2019}
T.~Nishio, H.~Okamoto, K.~Nakashima, Y.~Koda, K.~Yamamoto, M.~Morikura,
  Y.~Asai, and R.~Miyatake, ``Proactive received power prediction using machine
  learning and depth images for {mmWave} networks,'' \emph{{IEEE} J. Sel. Areas
  Commun.}, vol.~37, no.~11, pp. 2413--2427, 2019.

\bibitem{Koda2020}
Y.~Koda, J.~Park, M.~Bennis, K.~Yamamoto, T.~Nishio, M.~Morikura, and
  K.~Nakashima, ``Communication-efficient multimodal split learning for
  {mmWave} received power prediction,'' \emph{{IEEE} Commun. Lett.}, vol.~24,
  no.~6, pp. 1284--1288, 2020.

\bibitem{Xu2023_2}
W.~Xu, F.~Gao, Y.~Zhang, C.~Pan, and G.~Liu, ``Multi-user matching and resource
  allocation in vision aided communications,'' \emph{{IEEE} Trans. Commun.},
  vol.~71, no.~8, pp. 4528--4543, 2023.

\bibitem{Xu2023}
W.~Xu, F.~Gao, X.~Tao, J.~Zhang, and A.~Alkhateeb, ``Computer vision aided
  mmwave beam alignment in {V2X} communications,'' \emph{{IEEE} Trans. Wireless
  Commun.}, vol.~22, no.~4, pp. 2699--2714, 2023.

\bibitem{Feng2024}
Y.~Feng, F.~Gao, X.~Tao, S.~Ma, and H.~V. Poor, ``Vision-aided ultra-reliable
  low-latency communications for smart factory,'' \emph{{IEEE} Trans. Commun.},
  pp. 1--1, 2024.

\bibitem{env}
S.~Imran, G.~Charan, and A.~Alkhateeb, ``Environment semantic aided
  communication: {A} real world demonstration for beam prediction,'' in
  \emph{Proc. IEEE Int. Conf. Commun. Workshops (ICC Workshops)}, 2023, pp.
  48--53.

\bibitem{feifei2023}
Y.~Yang, F.~Gao, X.~Tao, G.~Liu, and C.~Pan, ``Environment semantics aided
  wireless communications: {A} case study of {mmWave} beam prediction and
  blockage prediction,'' \emph{{IEEE} J. Sel. Areas Commun.}, vol.~41, no.~7,
  pp. 2025--2040, 2023.

\bibitem{feifeigao2023}
F.~Wen, W.~Xu, F.~Gao, C.~Pan, and G.~Liu, ``Vision aided environment semantics
  extraction and its application in {mmWave} beam selection,'' \emph{{IEEE}
  Commun. Lett.}, vol.~27, no.~7, pp. 1894--1898, 2023.

\bibitem{myy}
Y.~Ma \emph{et~al.}, ``Orthogonal delay-doppler division multiplexing
  modulation with tomlinson-harashima precoding,'' \emph{IEEE Trans. Commun.},
  pp. 1--13, 2024.

\bibitem{Jocher_Ultralytics_YOLO_2023}
\BIBentryALTinterwordspacing
G.~Jocher, A.~Chaurasia, and J.~Qiu, ``{Ultralytics YOLO},'' Jan. 2023.
  [Online]. Available: \url{https://github.com/ultralytics/ultralytics}
\BIBentrySTDinterwordspacing

\bibitem{talaat2023improved}
F.~M. Talaat and H.~ZainEldin, ``An improved fire detection approach based on
  {YOLO-v8} for smart cities,'' \emph{Neural Comput. Appl.}, vol.~35, no.~28,
  pp. 20\,939--20\,954, 2023.

\bibitem{COCO}
T.-Y. Lin, M.~Maire, S.~Belongie, J.~Hays, P.~Perona, D.~Ramanan,
  P.~Doll{\'a}r, and C.~L. Zitnick, ``Microsoft coco: Common objects in
  context,'' in \emph{Proc. European Conf. Comput. Vis.}, 2014, pp. 740--755.

\bibitem{he2015deep}
K.~He, X.~Zhang, S.~Ren, and J.~Sun, ``Deep residual learning for image
  recognition,'' 2015.

\bibitem{russakovsky2015imagenet}
O.~Russakovsky \emph{et~al.}, ``Imagenet large scale visual recognition
  challenge,'' \emph{Int. J. Comput. Vis.}, vol. 115, pp. 211--252, 2015.

\bibitem{deepsense}
A.~Alkhateeb, G.~Charan, T.~Osman, A.~Hredzak, J.~Morais, U.~Demirhan, and
  N.~Srinivas, ``Deepsense {6G}: A large-scale real-world multi-modal sensing
  and communication dataset,'' \emph{{IEEE} Commun. Mag.}, vol.~61, no.~9, pp.
  122--128, 2023.

\bibitem{Charan_DeepSense_2022a}
G.~Charan, T.~Osman, A.~Hredzak, N.~Thawdar, and A.~Alkhateeb,
  ``Vision-position multi-modal beam prediction using real millimeter wave
  datasets,'' in \emph{Proc. IEEE Wireless Commun. Netw. Conf. (WCNC)}, 2022,
  pp. 2727--2731.

\bibitem{AhmedAlkhateeb2022}
A.~Alkhateeb and G.~Charan, ``Computer vision aided blockage prediction in
  real-world millimeter wave deployments,'' in \emph{Proc. IEEE Globecom
  Workshops (GC Wkshps)}, {Rio de Janeiro, Brazil}, Dec. 2022, pp. 1711--1716.

\bibitem{Charan2024}
G.~Charan and A.~Alkhateeb, ``User identification: {A} key enabler for
  multi-user vision-aided communications,'' \emph{IEEE Open J. Commun.
  Society}, vol.~5, pp. 472--488, 2024.

\bibitem{kingma2014adam}
D.~P. Kingma and J.~Ba, ``Adam: A method for stochastic optimization,''
  \emph{arXiv preprint arXiv:1412.6980}, 2014.

\bibitem{labelme}
\BIBentryALTinterwordspacing
K.~Wada, ``{Labelme: Image Polygonal Annotation with Python}.'' [Online].
  Available: \url{https://github.com/wkentaro/labelme}
\BIBentrySTDinterwordspacing

\bibitem{X-AnyLabeling}
W.~Wang, ``Advanced auto labeling solution with added features,''
  \url{https://github.com/CVHub520/X-AnyLabeling}, CVHub, 2023.

\end{thebibliography}
\end{document}